\documentclass[lettersize,journal]{IEEEtran}
\usepackage{amsmath,amsfonts}
\usepackage{algorithmic}
\usepackage{algorithm}
\usepackage[caption=false,font=normalsize,labelfont=sf,textfont=sf]{subfig}
\usepackage{textcomp}
\usepackage{stfloats}
\usepackage{url}
\usepackage{verbatim}
\usepackage{graphicx}
\usepackage{cite}
\usepackage{amssymb}
\usepackage{booktabs}
\graphicspath{{./figs/}}
\graphicspath{{./figs/webuav3m/}}
\DeclareGraphicsExtensions{.pdf,.jpg,.png}
\usepackage{cases}
\usepackage{mdwmath}
\usepackage[dvipsnames]{xcolor}
\usepackage{array}
\usepackage{threeparttable}
\usepackage{color}
\usepackage{wasysym}
\usepackage{makecell}
\usepackage{dsfont}
\usepackage{bbm}
\usepackage{mathrsfs}
\usepackage{pifont}
\usepackage{marvosym}
\usepackage{colortbl}  %
\usepackage{xcolor}
\usepackage{multirow,diagbox}  %
\usepackage{float}
\usepackage{subfig}
\usepackage{caption}

\usepackage{amsthm,amsmath,amssymb}

\usepackage{mathrsfs}
\usepackage{verbatim}

\newcommand{\myPara}[1]{\vspace{0.02in}\noindent\textbf{#1}}

\def\ie{{\em i.e.}}
\def\eg{{\em e.g.}}
\def\etal{{\em et al.}}

\makeatletter

\newcommand{\Rmnum}[1]{\expandafter\@slowromancap\romannumeral #1@}
\makeatother

\usepackage{tikz}
\usepackage{verbatim}
\usetikzlibrary{backgrounds,mindmap,trees}
\usetikzlibrary{calc,positioning,intersections}
\usepackage{pgfplots}
\usepackage{listings}

\usepackage{genealogytree}

\usepackage{mathrsfs}
\usepackage{multicol} 

\usepackage{hyperref}
\hypersetup{
    colorlinks=true,             %link color
    linkcolor=black,             %internal link color
    filecolor=black,             %file color
    urlcolor=black,              %url color
    citecolor=blue,              %reference color
    pdftitle={Overleaf Example},
    pdfpagemode=FullScreen,
}

\begin{document}

\title{Segment Anything for Videos: A Systematic Survey}

\author{Chunhui~Zhang,\IEEEmembership{~Student~Member,~IEEE},
        Yawen~Cui,
        Weilin~Lin,
        Guanjie~Huang,
        Yan~Rong,\\
        Li~Liu,\IEEEmembership{~Member,~IEEE},
        Shiguang Shan,~\IEEEmembership{Fellow, IEEE}
        % <-this % stops a space
\thanks{C.~Zhang is with Shanghai Jiao Tong University, The Hong Kong University of Science and Technology (Guangzhou), and also with the CloudWalk Technology Co., Ltd. E-mail: chunhui.zhang@sjtu.edu.cn.}

\thanks{Y.~Cui is with the University of Oulu, Finland, and also with The Hong Kong University of Science and Technology (Guangzhou), China. E-mail: yawencui@oulu.fi.}

\thanks{W.~Lin, G.~Huang, Y.~Rong, and L.~Liu are with The Hong Kong University of Science and Technology (Guangzhou). E-mails: \{wlin760, ghuang565, yrong854\}@connect.hkust-gz.edu.cn, avrillliu@hkust-gz.edu.cn.}

\thanks{S.~Shan is with the Key Lab of Intelligent Information Processing of Chinese Academy of Sciences (CAS), Institute of Computing Technology, CAS, and also with University of Chinese Academy of Sciences. E-mail: sgshan@ict.ac.cn.}

\thanks{C.~Zhang, Y.~Cui, W.~Lin, G.~Huang, and Y.~Rong contributed equally to this work.}
\thanks{This work was done at The Hong Kong University of Science and Technology (Guangzhou).}
\thanks{Li~Liu is the corresponding author.}
%\thanks{{\normalsize \Letter}~Li~Liu is the corresponding author (avrillliu@hkust-gz.edu.cn).}
}

\markboth{Journal of \LaTeX\ Class Files,~Vol.~XX, No.~XX, August~2024}%
{Shell \MakeLowercase{\textit{et al.}}: A Sample Article Using IEEEtran.cls for IEEE Journals}


\maketitle

\begin{abstract}
The recent wave of foundation models has witnessed tremendous success in computer vision (CV) and beyond, with the segment anything model (SAM) having sparked a passion for exploring task-agnostic visual foundation models. Empowered by its remarkable zero-shot generalization, SAM is currently challenging numerous traditional paradigms in CV, delivering extraordinary performance not only in various image segmentation and multi-modal
segmentation (\eg, text-to-mask) tasks, but also in the video
domain. Additionally, the latest released SAM 2 is once again sparking research enthusiasm in the realm of promptable visual segmentation for both images and videos. However, existing surveys mainly focus on SAM in various image processing tasks, a comprehensive and in-depth review in the video domain is notably absent. To address this gap, this work conducts a systematic review on SAM for videos in the era of foundation models. As the first to review the progress of SAM for videos, this work focuses on its applications to various tasks by discussing its recent advances, and innovation opportunities of developing foundation models on broad applications. We begin with a brief introduction to the background of SAM and video-related research domains. Subsequently, we present a systematic taxonomy that categorizes existing methods into three key areas: video understanding, video generation, and video editing, analyzing and summarizing their advantages and limitations. Furthermore, comparative results of SAM-based and current state-of-the-art methods on representative benchmarks, as well as insightful analysis are offered. Finally, we discuss the challenges faced by current research and envision several future research directions in the field of SAM for video and beyond.
\end{abstract}


%------------------------------------------------------------------------------

\section{Introduction}

\IEEEPARstart{F}{oundation} models~\cite{bommasani2021opportunities,wang2023large,chunhui2023samsurvey} have become a significant area of research in recent years, revolutionizing various fields such as natural language processing (NLP), computer vision (CV), and machine learning. These models are typically pre-trained on massive datasets, enabling them to learn general representations of the input data and extract meaningful features that can be further fine-tuned for specific applications. While foundation models have primarily garnered extensive attention in NLP, their utility extends beyond that domain. In CV, researchers have been exploring the application of foundation models to enhance imaging understanding~\cite{chen2023sam,kim2023medivista,shi2023flowformer++}, object detection~\cite{liu2023grounding,lin2015microsoft}, image segmentation~\cite{zhang2023segment,oquab2023dinov2}, and other vision-related tasks~\cite{Non-euclidean_segment,zhang2023meta}. 

\begin{figure}[t]
\centering
\includegraphics[width =1.0\columnwidth]{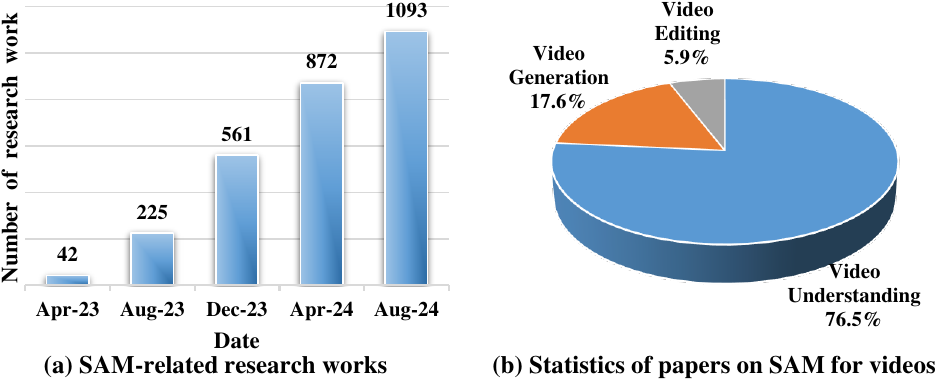}

\caption{Summarization on SAM-based works. (a) The number of SAM-related research works is rapidly increasing. (b) Video understanding dominates the research of SAM for videos.}
\label{fig:StatisticofSAMbasedworks}

\end{figure}

One prominent example is the segment anything model (SAM)~\cite{ICCV2023SAM}, which has achieved remarkable progress in exploring general and task-agnostic foundation models in the CV community. By training on over 1 billion masks on 11 million images, SAM can deliver high-quality segmentation masks based on multiple prompts (\eg, points, box, and text). More importantly, SAM exhibits powerful zero-shot generalization in various segmentation tasks (\eg, interactive segmentation, semantic segmentation, and panoptic segmentation), without the retraining or finetuning previously required~\cite{liu2023matcher}. Therefore, the emergence of SAM has led many researchers to believe that this is \emph{``the GPT-3 moment for CV, as SAM has learned the general concept of what an object is, even for unknown objects, unfamiliar scenes (\eg, underwater and cell microscopy and ambiguous cases)''}~\cite{GPT3moment}. A large number of researchers have extended SAM to different fields~\cite{ke2023segment,mo2023av,wang2023detect,zhang2023personalize,zhang2023uvosam}. As shown in Fig.~\ref{fig:StatisticofSAMbasedworks}(a), the number of SAM-related research works has increased significantly since April 2023\footnote{https://github.com/liliu-avril/Awesome-Segment-Anything}. The segment anything model 2 (SAM 2)~\cite{ravi2024sam2} enhances its predecessor, SAM, by integrating a transformer framework with streaming memory, facilitating superior real-time video segmentation capabilities. Trained on the extensive and diverse segment anything video (SA-V) dataset, SAM 2 demonstrates heightened accuracy and efficiency over SAM, particularly in video tasks, and offers a robust solution for promptable visual segmentation across varied spatio-temporal contexts.

\myPara{Incorporating SAM into Video Tasks.} Video is an incredibly important medium in today's digital age~\cite{xing2023videodiffusion}. Compared to static image and pure text, video offers strong visual representation, enhanced perception and memory, powerful storytelling capabilities, and rich interactivity, making it a more effective medium for communication and entertainment~\cite{xing2023videodiffusion,wang2023large}. The exploration of SAM in video tasks is quickly becoming a booming area of research~\cite{chunhui2023samsurvey,yang2023track,videoTextSpottingSA,Lu2023CanSB}. Although SAM has shown great potential in various image tasks, it still faces numerous challenges in video tasks, such as ensuring SAM's ability to consistently and coherently generate masks across lengthy video frames~\cite{zhang2023personalize,ke2023segment} and enhancing its scalability and efficiency for handling large-scale video data~\cite{yao2023sam, cheng2023segment}. \textcolor{black}{Most current works on video-related tasks usually employ SAM directly to achieve remarkable results of innovative applications. For a comprehensive understanding of this cutting-edge research field, as illustrated in Fig.~\ref{fig:StatisticofSAMbasedworks}(b), we conducted this survey and categorized existing works of innovative applications with SAM into three major categories (\ie, video understanding, video generation, and video editing)}.

\myPara{Unique Challenges in Videos.} Compared with other tasks, \eg, image and text processing, video tasks present the following unique challenges~\cite{chunhui2023samsurvey,videoTextSpottingSA,yang2023track,xing2023videodiffusion,Lu2023CanSB}. \textbf{1)} Temporal information processing: video data encompasses not only spatial information but also temporal dynamics. Thus, handling video data requires considering the temporal relationships and dynamic changes. \textbf{2)} High-dimensional data: each frame of a video consists of high-dimensional data with a large number of pixels, leading to a massive amount of data that demands more computational resources and storage space. \textbf{3)} Continuity and stability: videos are generally continuous, and processing them involves considering the coherence and stability between frames to achieve reliable results in analysis and applications. \textbf{4)} Time cost: due to the substantial volume of video data, the time cost for processing video tasks is usually higher, posing greater demands on computational resources and algorithm efficiency. \textbf{5)} Action and event recognition: compared to static images, video tasks often involve recognizing actions and events, requiring models to understand and learn dynamic changes in temporal sequences. The above challenges foreshadow the extreme complexity of video tasks and enormous research opportunities~\cite{xing2023videodiffusion,cheng2023segment,ke2023segment}.

%\begin{figure}[t]
%\centering
%\includegraphics[width =1.0\columnwidth]{figs/SAM_Framework.pdf}

%\caption{Dataflow of SAM. According to the user prompt, SAM can achieve interactive segmentation. Several research routes for the SAM model (\eg, model compression~\cite{zhao2023fast}, model robustness~\cite{zhang2023attack}), prompt (\eg, efficient finetuning~\cite{chen2023sam}), and outputs (\eg, innovative applications~\cite{zhang2023segment,lai2023detect}) are listed.}
%\label{fig:SAM}
%\end{figure}

\begin{figure*}[t]
\centering
\includegraphics[width=1.0\linewidth]{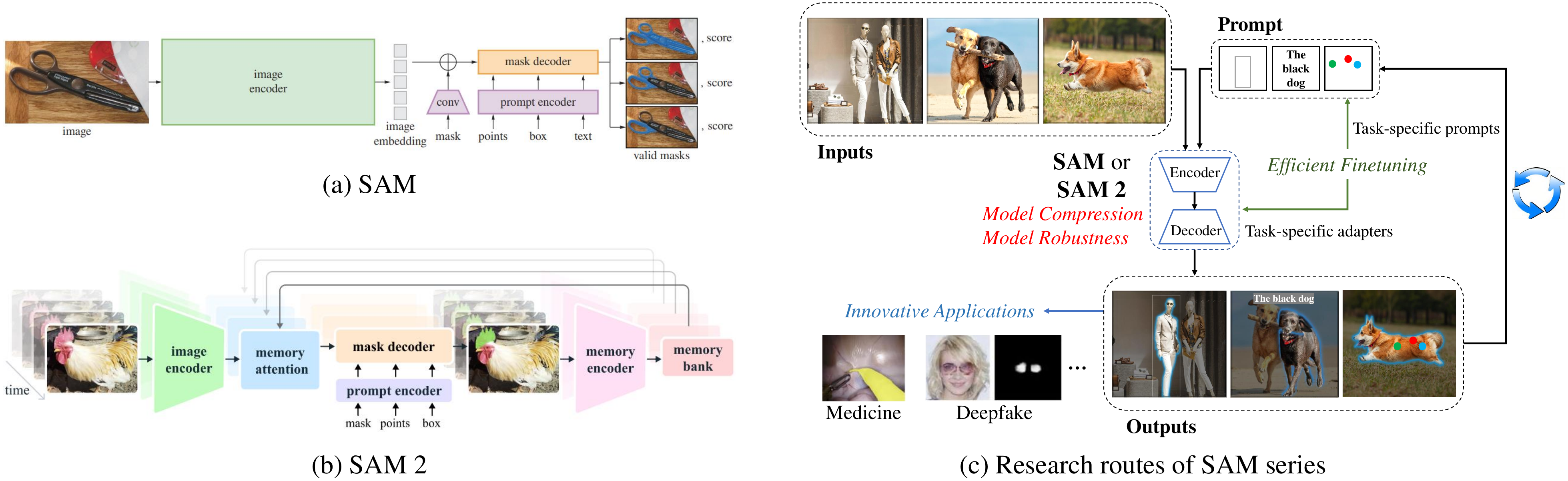}

\caption{Overall architectures of SAM (a) and SAM 2 (b) from the original papers~\cite{ICCV2023SAM,ravi2024sam2}, respectively. According to the user prompts, SAM and SAM 2 can achieve interactive segmentation in images and videos. Several representative research routes for the SAM and SAM 2 models (\eg, model compression~\cite{zhao2023fast}, model robustness~\cite{zhang2023attack}), prompt (\eg, efficient finetuning~\cite{chen2023sam}), and outputs (\eg, innovative applications~\cite{zhang2023segment,lai2023detect}) are listed in (c).}
\label{fig:SAM}
\end{figure*}

\myPara{Comparisons with Previous Surveys.}
Although three surveys~\cite{chunhui2023samsurvey,zhang2023survey,zhang2023segment} have been proposed for SAM, the differences between our survey and existing ones are mainly in three aspects. \textbf{1)} Previous SAM-based surveys only focus on medical image segmentation tasks~\cite{zhang2023segment} or roughly cover video tasks~\cite{chunhui2023samsurvey,zhang2023survey}, however, SAM for videos is a challenging and promising research topic with many innovation opportunities and potential applications~\cite{xing2023videodiffusion}. This inspires us to conduct a systematic survey dedicated to this specific field (\ie, SAM for videos) to benefit relevant researchers and practitioners. \textbf{2)} This survey provides an understandable and highly structured taxonomy of SAM for videos, dividing existing methods into three major categories (\ie, video understanding, video generation, and video editing), which is significantly different from previous ones. \textbf{3)} A comprehensive performance evaluation, together with many new insights on SAM for videos are offered to help readers track recent advances. Additionally, the proposed research directions are deliberate and can pave new avenues for developing foundation models in the video domain and beyond. For a comprehensive understanding of foundation models, we also refer readers to other excellent surveys for language~\cite{mialon2023augmented,fan2023bibliometric,blodgett2020language}, vision~\cite{long2022vision,xing2023videodiffusion}, and multi-modality~\cite{bommasani2021opportunities,wang2023large}.

The main contributions of this survey are threefold:
\begin{itemize}
\item{We thoroughly review the development of SAM for videos in the foundation models era and provide a systematic survey of the latest progress in this field, which can be grouped into three major categories: video understanding, video generation, and video editing. To the best of our knowledge, this is the first systematic survey that focuses on this specific domain.}

\item{We comprehensively compare SAM-based methods with current state-of-the-art (SOTA) methods on representative datasets for various video tasks. Importantly, our in-depth analysis about the pros and cons of these leading-edge methods can help readers choose appropriate baselines for their specific applications while delivering valuable insights on improving existing methods.}

\item{Based on the systematic literature review and comprehensive performance evaluation, we highlight some potential future developmental trends.}
\end{itemize}

The remainder of this survey is organized as follows. Section~\ref{sec:Preliminaries} summarizes the background knowledge, including the workflows of SAM and SAM 2, research routes, and relevant research domains. In Section~\ref{sec:class_1}, we primarily present an overview of methods in the field of video understanding with SAM. In Section~\ref{sec:class_2}, we delve into the principal studies concerning video generation with SAM. In Section~\ref{sec:class_3}, we elucidate the methods for video editing with SAM. Section~\ref{sec:PerformanceEvaluation} introduces the benchmark datasets and evaluation. In Section~\ref{sec:Outlooks}, we conclude this article and highlight the potential avenues for future research.

%------------------------------------------------------------------------------
\section{Preliminaries}
\label{sec:Preliminaries}

In this section, we first briefly introduce SAM, then review three video-related research domains, including video understanding, video generation, and video editing. 

\subsection{Segment Anything Models}
SAM is the segment foundation model proposed by Meta \cite{ICCV2023SAM}, as illustrated in Fig.~\ref{fig:SAM}(a). The pathway of SAM consists of three steps, namely task, model, and data. Inspired by large language models, tasks in SAM are usually introduced using prompt engineering \cite{brown2020language}, where a prompt is to indicate what to segment. A unique characteristic of the promptable task is that it can return a valid segmentation mask when given any segmentation prompt. The structure of SAM consists of three parts: a powerful image encoder (\textit{i.e.}, ViT \cite{dosovitskiy2021image}); a prompt encoder, dense input, and a mask decoder (prompt-image bidirectional Transformer decoder using self-attention and cross-attention). The model is trained with focal loss \cite{lin2018focal} and dice loss \cite{milletari2016vnet}. Due to the insufficiency of public training data for segmentation tasks, the training-annotation iterative process is conducted in SAM by constructing a data engine to achieve model training and dataset construction simultaneously. Benefiting from well-designed tasks, the model structure, and an extensive repository of high-quality training data, experiments demonstrate that the SAM model excels in zero-shot transfer capabilities. It has shown remarkable performance in tasks such as single-cue point segmentation, edge detection, object proposal, instance segmentation, interactive segmentation, and multi-modal segmentation (text-to-mask). Notably, the SAM model even surpasses supervised models in certain aspects.

The latest SAM 2~\cite{ravi2024sam2} (see Fig.~\ref{fig:SAM}(b)) introduces a significant evolution over its predecessor by extending its capabilities to the domain of video segmentation. SAM 2 incorporates a transformer-based architecture with a streaming memory component, enabling real-time processing of video frames. It refines the segmentation process through interactive user prompts and leverages a memory attention mechanism to retain and utilize information about the target object across frames. The SAM 2 model demonstrates improved accuracy and efficiency, requiring fewer interactions for video segmentation and outperforming SAM in both speed and accuracy for image segmentation tasks. Furthermore, the SAM 2 model is trained on the SA-V dataset, which is a substantial expansion from SAM's training data. The SA-V dataset, comprising 50.9K videos with 642.6K masklets, is not only larger but also more diverse, covering a wider range of objects and scenarios. This extensive and varied dataset has been instrumental in enhancing SAM 2's ability to segment objects in complex, real-world video content, thereby setting a new benchmark for visual segmentation tasks. The improvements in SAM 2 reflect a concerted effort to address the dynamic challenges present in video data, such as motion, deformation, and occlusion, and to provide a more generalized solution for promptable visual segmentation.

\subsection{Research Routes of SAM} 
%Research on SAM mainly adapts the following routes: model compression~\cite{zhao2023fast}, model robustness~\cite{zhang2023attack}, efficient finetuning~\cite{chen2023sam}, and innovative applications~\cite{zhang2023segment,lai2023detect}, from methodology (see Fig.~\ref{fig:SAM}(c)). For video-related tasks, most of works with SAM belong to innovative applications, which directly use SAM to obtain remarkable outputs. Other works mainly refer to efficient finetuning to specific tasks. In this way, the taxonomy in this paper is based on innovative applications across various video tasks.

Research on SAM mainly adapts the following routes: model compression~\cite{zhao2023fast}, ensuring model robustness~\cite{zhang2023attack}, advancing efficient finetuning techniques~\cite{chen2023sam}, and developing innovative applications~\cite{zhang2023segment,lai2023detect} (as illustrated in Fig.~\ref{fig:SAM}(c)) from the perspective of methodology. In the realm of video processing, the majority of SAM research falls under the category of innovative applications, where SAM is directly applied to achieve significant outcomes. Meanwhile, a portion of the research is dedicated to refining finetuning approaches tailored for individual video tasks. Consequently, the taxonomy in this paper is based on innovative applications that SAM enables across diverse video-related challenges.

\subsection{Related Tasks}
\myPara{Video Understanding.} 
Video understanding aims to recognize and localize different actions or events appearing in the video, including (1) video recognition and (2) video localization. (1) Video recognition aims to classify the video clip or snippet into one of action or event categories. Frameworks of current works are mainly divided into two series: two-stream networks~\cite{9869674,10155270} and single-stream RGB networks~\cite{carreira2017quo,feichtenhofer2019slowfast}. This work~\cite{simonyan2014two} proposes a two-stream ConvNet architecture which incorporates spatial and temporal networks and demonstrates that a ConvNet trained on multi-frame dense optical flow is able to achieve very good performance in spite of limited training data. SlowFast networks~\cite{feichtenhofer2019slowfast}, a one-stream framework, consists of a fast pathway operating at high frame rate and a slow pathway operating at low frame rate. (2) Video localization targets to detect and classify actions in untrimmed long videos. There are two widely used detection paradigms. The two-stage
paradigm~\cite{7236896,tan2021relaxed} first localizes class-agnostic action proposal, then classifies and refines
each proposal. Another one-stage paradigm~\cite{nawhal2021activity,zhang2022actionformer} combines localization and classification, which densely classifies each frame into actions or backgrounds.

\myPara{Video Generation.} Video generation aims to generate new videos from (1) the text (\textit{i.e.}, text-to-video generation) or from (2) a single video. (1) Text-to-video generation. Early works~\cite{9439899,964787,zhang2022actionformer} primarily generate videos in simple domains, such as moving digits or specific human actions. Recently, a series of works~\cite{van2017neural, wu2022nuwa} conduct VAE-based methods for more realistic scenes. Inspired by text-to-image diffusion models, Video
Diffusion Models (VDM)~\cite{ho2022imagen} are proposed with a space-time factorized U-Net with joint image and video data training. Make-A-Video~\cite{singer2022make} and MagicVideo~\cite{zhou2022magicvideo} aim to generate videos by transferring progress from text-to-image generation. (2) Video generation from a single video. The methods on this task are divided into GAN-based methods~\cite{arora2021singan, gur2020hierarchical} and Patch nearest-neighbour methods~\cite{haim2022diverse}. Sinfusion~\cite{nikankin2022sinfusion} is the first work to utilize the capabilities of diffusion models to learn the appearance and dynamics of a single video for generating new videos.

\myPara{Video Editing.} Video editing usually refers to editing a video according to the textual information or an example. Video stylization is a specific type of editing task where the style provided by an example frame is propagated to the video. Existing methods can be roughly divided into (1) propagation-based methods and (2) video layering-based methods. (1) Propagation-based methods use keyframes~\cite{jamrivska2019stylizing, texler2020interactive} to propagate edits throughout the video. This work~\cite{jamrivska2019stylizing} proposes a new type of guidance for SOTA patch-based synthesis, which can be applied to any type of video content. (2) Layer-based methods~\cite{kasten2021layered, lu2021omnimatte} usually decompose the video into layers that are then edited. Layered neural atlases~\cite{kasten2021layered} map the foreground and background of a video to a canonical space, which is then operated for video editing.

\begin{figure*}[htbp]
\centering
\begin{tikzpicture}[scale=1.0]  
\genealogytree[
%level 4/.style={level size=4.4cm,node box={colback=black!30}},
%level 3/.style={level size=4.5cm,node box={colback=black!30}},
%level 2/.style={level size=2.5cm,node box={colback=black!30}},
%level 1/.style={level size=2.4cm,node box={colback=red!30}},
%level 0/.style={level size=2.0cm,node box={colback=yellow!30}},
level 4/.style={level size=4.4cm,node box={colback=black!30}},
level 3/.style={level size=4.5cm,node box={colback=black!30}},
level 2/.style={level size=2.5cm,node box={colback=black!30}},
level 1/.style={level size=2.4cm,node box={colback=red!30}},
level 0/.style={level size=2.0cm,node box={colback=yellow!30}},
timeflow=left,
processing=tcolorbox,
node size from=5mm to 4cm,
tcbset={male/.style={colframe=black},female/.style={colframe=black,colback=yellow!20}},
%box={size=small,halign=center,valign=center,fontupper=\small\sffamily},
box={size=small,halign=center,valign=center,fontupper=\footnotesize\sffamily},
highlight/.style={pivot,box={colback=gray!20,fuzzy halo}},
edges={foreground=black!25,background=black!5},
class1/.style={box={colback=red!10}},
class2/.style={box={colback=green!20}},
class3/.style={box={colback=blue!20}},
]{
    parent{
        g[female,box={width=2.0cm}]{SAM for Videos}
        parent{
            g[class1,male]{Video Understanding ($\S$ \ref{sec:class_1})}
            parent
            {
                g[class1,male]{Video Object Segmentation}
                parent{
                    g[class1,male]{Video Semantic Segmentation}
                    p[class1,male]{PerSAM~\cite{zhang2023personalize}, Matcher~\cite{liu2023matcher}, UVOSAM~\cite{zhang2023uvosam}, SSA~\cite{chang20233rd}, DSEC-MOS~\cite{zhou2023dsec}}
                    }
                parent{
                    g[class1,male]{Video Instance Segmentation}
                    p[class1,male]{HQ-SAM~\cite{ke2023segment}}
                    }
                parent{
                    g[class1,male]{Video Panoptic Segmentation}
                    p[class1,male]{DEVA~\cite{cheng2023tracking}}
                    }
                parent{
                    g[class1,male]{Video Entity Segmentation}
                    p[class1,male]{EntitySeg~\cite{qi2023high}}
                    }
            }  
            parent{
                g[class1,male]{Video Object Tracking}
                parent{
                    g[class1,male]{General Object Tracking}
                    p[class1,male]{TAM~\cite{yang2023track}, SAM-Track~\cite{cheng2023segment}, HQTrack~\cite{zhu2023tracking}}
                    }
                parent{
                    g[class1,male]{Open-Vocabulary Tracking}
                    p[class1,male]{OVTracktor~\cite{chu2023zero}}
                    }    
                parent{
                    g[class1,male]{Point Tracking}
                    p[class1,male]{SAM-PT~\cite{rajivc2023segment}}
                    }  
                parent{
                    g[class1,male]{Nighttime UAV Tracking}
                    p[class1,male]{SAM-DA~\cite{yao2023sam}}
                    } 
                }
            parent{
                g[class1,male]{Deepfake Detection}
                p[class1,male]{DADF~\cite{lai2023detect}}
                }
            parent{
                g[class1,male]{Video Shadow Detection}
                p[class1,male]{ShadowSAM~\cite{wang2023detect}}
                }
            parent{
                g[class1,male]{Miscellaneous}
                parent{
                    g[class1,male]{Audio-Visual Segmentation}
                    p[class1,male]{AV-SAM~\cite{mo2023av}, GAVS~\cite{wang2023prompting}, CMSF~\cite{bhosale2023leveraging}}
                    }
                parent{
                    g[class1,male]{Referring Video Object Segmentation}
                    p[class1,male]{RefSAM~\cite{li2023refsam}}
                    }                
                }

            parent{
                g[class1,male]{Domain Specific}
                parent{
                    g[class1,male]{Medical Videos}
                    p[class1,male]{SurgicalSAM~\cite{yue2023surgicalsam}, SAMSNeRF~\cite{lou2023samsnerf}, MediViSTA-SAM~\cite{kim2023medivista}, SuPerPM~\cite{lin2023superpm}}
                    }
                parent{
                    g[class1,male]{Domain Adaptation}
                    p[class1,male]{L-SAM~\cite{bonani2023learning}, SAM-DA~\cite{yao2023sam}}
                    }
                parent{
                    g[class1,male]{Tool Software}
                    p[class1,male]{TD~\cite{hsieh2023tool}}
                    }
                parent{
                    g[class1,male]{More Directions}
                    p[class1,male]{SAMFlow~\cite{zhou2023samflow}, MMPM~\cite{yang2023pave}, SemCom~\cite{raha2023generative}, SAM-RL~\cite{schiller2023virtual}, ROSGPT\_Vision~\cite{benjdira2023rosgpt_vision}}
                    }
                }
                
            }
            
        parent{
                g[class2,male]{Video Generation ($\S$ \ref{sec:class_2})}
                parent{
                    g[class2,male]{Video Synthesis}
                    p[class2,male]{Dancing Avatar~\cite{qin2023dancing},  DISCO~\cite{wang2023disco}}
                    }
               parent{
                    g[class2,male]{Video Super-Resolution}
                    p[class2,male]{SEEM~\cite{Lu2023CanSB}}
                    }
               parent{
                    g[class2,male]{3D Reconstruction}
                    p[class2,male]{SAM3D~\cite{yang2023sam3d}, CSF~\cite{dong2023leveraging}, OSTRA~\cite{xu2023one}}
                    }
               parent{
                    g[class2,male]{Video Dataset Annotation Generation}
                    p[class2,male]{SLP~\cite{balaban2023propagating}, SAMText~\cite{videoTextSpottingSA}, AVISeg~\cite{guo2023audio}, EVA-VOS~\cite{delatolas2023learning}, BOArienT \& SeaDronesSee-3D~\cite{kiefer2023stable}}
                    }
            }

        parent{
                g[class3,male]{Video Editing ($\S$ \ref{sec:class_3})}
               parent{
                    g[class3,male]{Generic Video Editing}
                    p[class3,male]{Make-A-Protagonist~\cite{zhao2023make}}
                    }
               parent{
                    g[class3,male]{Text Guided Video Editing}
                    p[class3,male]{2SVE~\cite{wu2023cvpr}}
                    }
               parent{
                    g[class3,male]{Object Removing}
                    p[class3,male]{OR-NeRF~\cite{yin2023or}}
                    }
            }
            
        }
    }
\end{tikzpicture}
\caption{Taxonomy of research works on SAM for videos. Due to space considerations, we merely list some representative methods for each video-related task here.}
\label{fig:sam_based_methods}

\end{figure*}
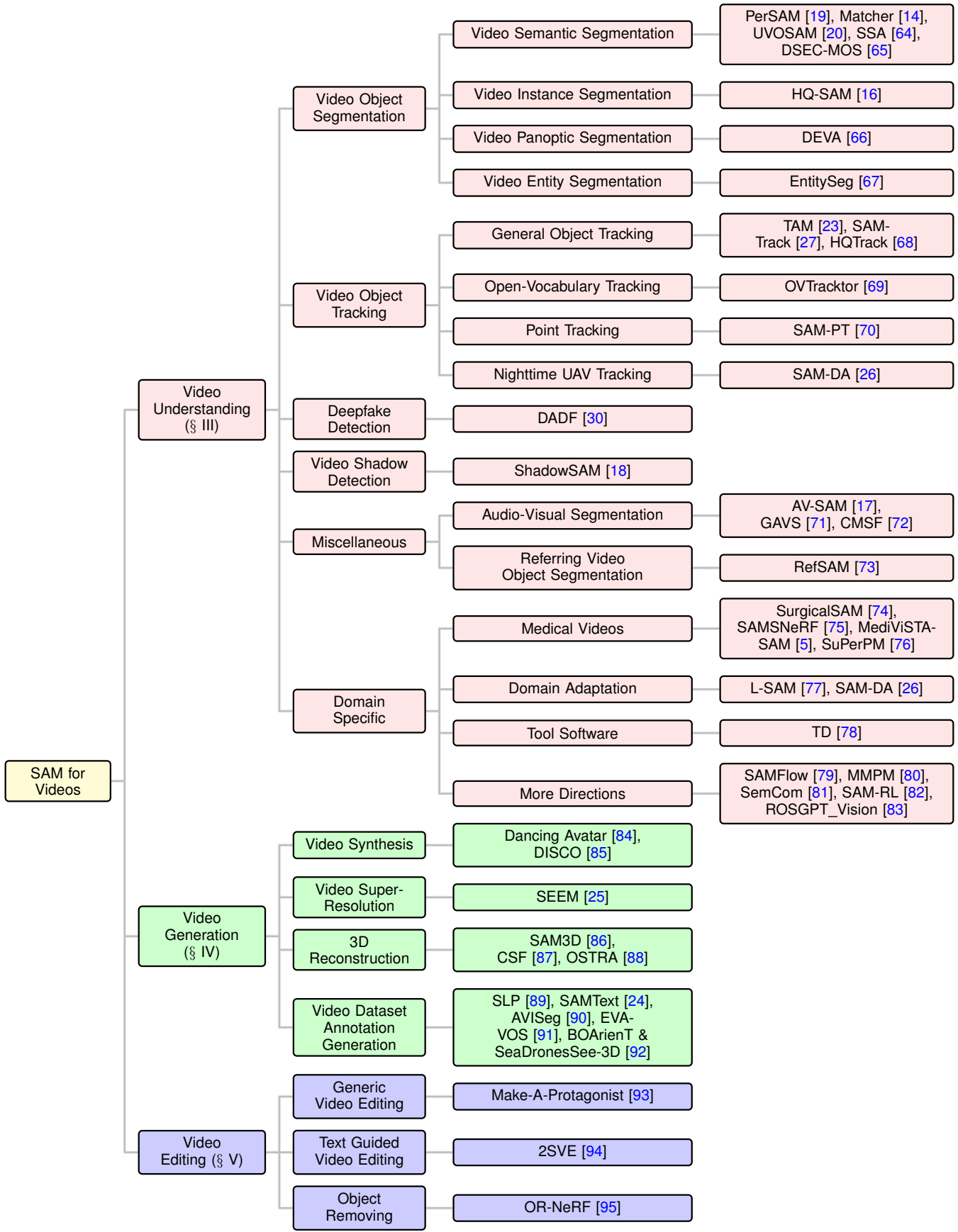

%------------------------------------------------------------------------------
\section{Video Understanding with SAM}
\label{sec:class_1}
In this section, we primarily introduce various video understanding tasks using SAM, as shown in Fig.~\ref{fig:sam_based_methods}.

\subsection{Video Object Segmentation} 

\begin{figure*}[t]
\centering
\includegraphics[width =1.0\linewidth]{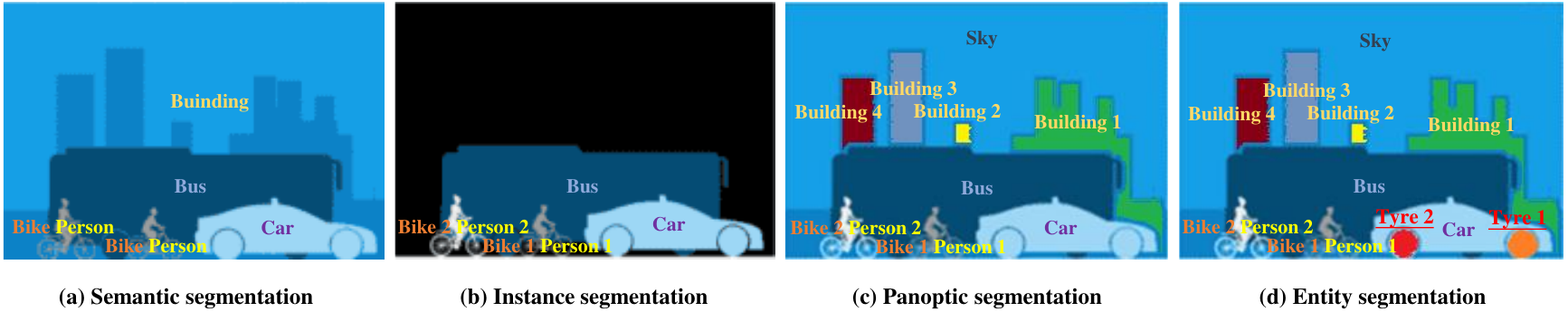}
\caption{Concepts comparison of four prevalent visual segmentation tasks, including semantic, instance, panoptic, and entity segmentation. (a) For semantic segmentation, the same textures or categories are assigned the same class labels. (b) Instance segmentation only focuses on the foreground, and different objects in the same category are assigned different instance identities. (c) In panoptic segmentation, each pixel is assigned a semantic label and a unique instance identifier. (d) Entity segmentation~\cite{qi2023high} requires segmenting unseen categories in the training set, \eg, ``tyre''.}
\label{fig:video_segmentation}
\end{figure*}

Video object segmentation (VOS) is a crucial task in CV for segmenting primary objects in a video. By combining with the pre-trained segmentation model SAM, recent works present great potential in VOS. We briefly summarize them into semantic, instance, panoptic, and entity levels (see Fig.~\ref{fig:video_segmentation}).

\subsubsection{Video Semantic Segmentation}
Zhang \etal \cite{zhang2023uvosam} was the first to adopt SAM for unsupervised VOS, which performs segmentation without manual annotations. Specifically, they remove the mask prediction branch in IDOL \cite{wu2022defense} to adapt it as a novel video salient object tracking method, which is to discover the salient object and spatial-temporal trajectories. Then, they adopt SAM with the generated trajectories as the prompt to obtain mask results frame by frame.

Besides, one-shot object segmentation customizing the great segmentation ability of SAM also works well in image segmentation and video segmentation. Liu \etal \cite{liu2023matcher} presents a training-free framework, Matcher, with one-shot object segmentation. They integrate an all-purpose feature extraction model (\eg, DINOv2~\cite{oquab2023dinov2}, CLIP~\cite{radford2021learning}, and MAE~\cite{he2021masked}) and a class-agnostic segmentation model (\ie, SAM) with three operations to realize the controllable masks generation. Following it, Zhang \etal \cite{zhang2023personalize} introduce a training-free personalization SAM, named PerSAM, to segment only the user-provided object with SAM. Specifically, they first obtain a location confidence map for the target object with the user-provided image and mask. Then, based on the confidence scores, they proposed target-guided attention and target-semantic prompting to aid SAM's decoder for personalized segmentation. Additionally, they provide a finetuning variant PerSAM-F with only 2 parameters within 10 seconds to alleviate the mask ambiguity issue. Both of the methods can be used in image and video object segmentation in the frame-by-frame setting.

Except for them, Chang \etal \cite{chang20233rd} adopting SAM as a post-processing technique for semantic segmentation in the PVUW2023 VSS track. Zhou \etal \cite{zhou2023dsec} propose a novel moving object segmentation (MOS) dataset, named DSEC-MOS, with high temporal resolution and low-latency information on the changes of scenes to promote the research on MOS.

\subsubsection{Video Instance Segmentation}
To solve the problems of coarse mask boundaries and incorrect predictions on SAM, Ke \etal \cite{ke2023segment} propose HQ-SAM, which equips SAM with the ability to segment any object more accurately. To be specific, they introduce a lightweight High-quality Output Token to replace the original SAM's output token and a Global-local Feature Fusion to fuse the global semantic context and the local boundary details. They fix the pre-trained model parameters to keep the original performance of SAM and only train a few parameters of the introduced components on their composed dataset with 44K fine-grained masks.

\subsubsection{Video Panoptic Segmentation}
One essential challenge for the end-to-end video segmentation model is poor performance in large-vocabulary settings. In the large-vocabulary dataset VIPSeg~\cite{miao2022large}, a recent work \cite{li2022video} achieves only 26.1 in terms of video panoptic quality score. Cheng \etal \cite{cheng2023tracking} state that the increasing number of classes and scenarios makes it difficult to conduct end-to-end training with good performance. Therefore, they propose a decoupled video segmentation approach (DEVA) with task-specific image-level segmentation and class/task-agnostic bi-directional temporal propagation. Specifically, SAM is used for image-level segmentation with universal data training containing outside-the-target-domain data. With the first segmented frame, they denoise the error with a few frames in the near future to reach a consensus as the output segmentation. Then, XMem~\cite{cheng2022xmem} is adapted as the temporal propagation model to propagate the segmentation to subsequent frames. The extensive experiments on VIPSeg validate its effectiveness on large-scale video panoptic segmentation.

\subsubsection{Video Entity Segmentation}
The in-the-wild setting of the image/video segmentation task is a big challenge for the existing methods, where no restriction is set on domains, classes, image resolution, and quality \cite{qi2023high}. Although entity segmentation is designed to segment unseen categories  in the training set, the lack of entity segmentation datasets makes it difficult to develop well on this task. To fill this gap, Qi \etal \cite{qi2023high} construct a high-quality large-scale entity segmentation dataset, named EntitySeg. The dataset contains 33,227 images with high-quality annotated masks on multiple domains and diverse resolutions, allowing the evaluation of the models' generalization and robustness. They benchmark the existing models and find that they cannot well accommodate the proposed dataset. Therefore, they further propose the CropFormer~\cite{qi2023high} framework to solve the problem.

\subsection{Video Object Tracking}

% Video object tracking (VOT) is a fundamental task in CV. Recently, the strong segmentation ability of SAM enhances the perception of objects and allows better development in VOT. Yang \etal \cite{yang2023track} propose the training-free track anything model (TAM) based on SAM to achieve high-performance interactive tracking and segmentation in videos. Specifically, they first use SAM to get an initial mask of objects, where the user can choose the target object by a click or modify the mask. Then, they adopt XMem to perform VOS on the following frames with the user-selected mask. To avoid the issue that XMem segments more coarsely over time, they use SAM again to refine it. During the tracking process, users are allowed to compulsively stop and correct the current mask. With this pipeline, TAM shows superior performance in VOT.

Video object tracking (VOT) is a fundamental task in CV. We divide VOT methods using SAM into four groups: (1) general object tracking, (2) open-vocabulary tracking, (3) point tracking, and (4) nighttime unmanned aerial vehicle (UAV) tracking. (1) Recently, the strong segmentation ability of SAM enhances the perception of objects and allows better development in general object tracking. Yang \etal \cite{yang2023track} propose the training-free track anything model (TAM) based on SAM to achieve high-performance interactive tracking and segmentation in videos. Specifically, they first use SAM to get an initial mask of objects, where the user can choose the target object by a click or modify the mask. Then, they adopt XMem to perform VOS on the following frames with the user-selected mask. To avoid the issue that XMem segments more coarsely over time, they use SAM again to refine it. Cheng \etal \cite{cheng2023segment} propose SAM-Track to segment and track any object in a video. They incorporate SAM to obtain segments, Grounding-DINO to understand natural language, and DeAOT~\cite{yang2022decoupling} for tracking. In the VOTS2023 challenge, Zhu~\etal~\cite{zhu2023tracking} won 2$^{nd}$ place to achieve high-quality VOT with their proposed HQTrack. Specifically, the framework implements the improved variants of DeAOT and SAM (\ie, HQ-SAM~\cite{ke2023segment}) for multi-object segmentation and mask refining, respectively. A similar idea of combining SAM and DeAOT is seen in the 1$^{st}$ place solution of TREK-150 object tracking challenge~\cite{xu2023zju}. They introduce MSDeAOT as an improved variant of DeAOT by replacing the bounding box with masks in the reference frame and feeding the mask and frames into the VOS model. (2) Chu~\etal~\cite{chu2023zero} utilize SAM as the segmenter along with an open-vocabulary object detector and an optical flow estimation to build a zero-shot open-vocabulary visual tracking framework OVTracktor. 
(3) SAM-PT~\cite{rajivc2023segment} was proposed to utilize the sparse point propagation of VOS. Taking a video with point annotations in the first frame as input, SAM-PT can achieve strong zero-shot performance with a point tracker to generate the trajectories as prompts and SAM to output predicted masks. The predicted masks are also used to reinitialize and get rid of the unreliable points. (4) Yao~\etal~\cite{yao2023sam} utilize SAM for the field of real-time nighttime UAV tracking to accurately locate the potential object and determine high-quality target domain training samples from the night-time images.

\subsection{Deepfake Detection}
In a recent investigation, Lai \textit{et al.} \cite{lai2023detect} delved into evaluating the performance of SAM and its variants in the context of deepfake detection and localization, marking the first attempt to assess these methods for this specific task. The researchers noted that existing approaches, which utilize LoRA~\cite{zhou2014low}, SAM adapter \cite{chen2023sam}, and learnable prompt \cite{qiu2023learnable} to fine-tune SAM on downstream tasks, often yielded unsatisfactory results, particularly in terms of face forgery localization. This inadequacy was attributed to their limited capacity in modeling both local and global contexts for forgery.

To tackle these challenges, Lai \textit{et al.} \cite{lai2023detect} proposed an innovative framework, named detect any deepfakes (DADF), building upon SAM. Specifically, they introduced a multi-scale adapter within SAM designed to capture short- and long-range forgery contexts, facilitating efficient finetuning. Additionally, a reconstruction guided attention module was introduced to enhance forged traces and boost the model's sensitivity toward forgery regions. The proposed method exhibited SOTA performance in both forgery detection and localization.

\subsection{Video Shadow Detection}
The detection of video shadows plays a crucial role in various applications, including object detection \cite{cucchiara2003detecting}, image segmentation \cite{ecins2014shadow}, and virtual reality scene generation \cite{liu2020arshadowgan}. However, the challenge lies in the limited availability of training data, posing difficulties for the generalization capability of existing deep neural network based methods. These limitations can lead to prediction errors accumulating during video propagation \cite{xu2022reliable}.

In particular, when applying SAM to single-frame shadow detection, SAM tends to categorize shadows as part of the background \cite{wang2023detect}. This introduces a nontrivial challenge in using SAM for shadow detection, as it requires bridging the gap between natural objects and complex shadows. To address this challenge, Wang \textit{et al.} \cite{wang2023detect} introduced ShadowSAM, a straightforward yet effective framework designed for finetuning SAM specifically for shadow detection. Additionally, by adopting a long and short-term attention mechanism, they extended its capabilities for efficient video shadow detection.

\subsection{Miscellaneous}

\subsubsection{Audio-Visual Segmentation}
Recently, SAM was applied in audio-visual localization and segmentation \cite{mo2023av,wang2023prompting}. Both studies focus on overcoming challenges associated with audio-visual localization and segmentation, particularly addressing the inherent misalignment between audio and various objects in the video.

In \cite{mo2023av}, the authors address this challenge by introducing AV-SAM, a method that learns audio-aligned visual features for each mask prompt from the video. This facilitates the guidance of mask generation in SAM through pixel-wise audio-visual fusion. The approach utilizes audio features and visual features from the pre-trained image encoder in SAM to aggregate cross-modal representations. Conversely, Wang \textit{et al.} \cite{wang2023prompting} present an encoder-prompt-decoder paradigm to tackle issues related to data scarcity and varying data distribution. Leveraging abundant knowledge from pre-trained models, they introduce a semantic-aware audio prompt to assist the visual foundation model in focusing on sounding objects. Simultaneously, this approach encourages the reduction of the semantic gap between visual and audio modalities. Furthermore, Bhosale \textit{et al.} \cite{bhosale2023leveraging} propose CMSF, a method leveraging audio cues to generate audio tags and subsequently proposing segmentation masks. These recent advancements underscore the versatility of SAM in addressing intricate tasks related to audio-visual processing.

\subsubsection{Referring Video Object Segmentation}
Despite SAM gaining widespread attention for its impressive performance in image segmentation, a study discussed in \cite{li2023refsam} highlights SAM's limitations in the realm of referring video object segmentation (RVOS). This limitation stems from the need for precise user interactive prompts and a constrained understanding of different modalities, such as language and vision.

In a concerted effort to effectively tailor SAM for RVOS in an end-to-end manner and fully unleash its potential for video segmentation and multi-modal fusion, Li \textit{et al.} conducted a groundbreaking study \cite{li2023refsam}. They delved into SAM's potential for RVOS by integrating multi-view information from diverse modalities and successive frames at different timestamps. The authors introduced RefSAM, a novel approach that utilizes lightweight modules and an efficient finetuning strategy to align and fuse language and vision features in an end-to-end learning fashion. Additionally, they designed a hierarchical dense attention module to exploit diverse levels of visual and textual features, thereby facilitating effective cross-modal segmentation of objects with varying sizes.

\subsection{Domain Specific}

\subsubsection{Medical Videos} 
SAM also contributes to the analysis of medical videos. Regarding the two problems with naive pipeline of SAM (\textit{i.e.}, the domain gap and the dependency on precise point or box locations), SurgicalSAM~\cite{yue2023surgicalsam} introduces a novel end-to-end efficient finetuning approach for SAM, and the objective is to seamlessly incorporate surgical-specific information with SAM's pre-trained knowledge for enhancing overall generalization capabilities. This work~\cite{wang2023sam} comprehensively explores different scenarios of robotic surgery and evaluates SAM’s robustness and zero-shot generalizability. SAMSNeRF~\cite{lou2023samsnerf} combines SAM and neural radiance field (NeRF) techniques, which generates accurate segmentation masks of surgical tools using SAM and then guides the refinement of the dynamic surgical scene reconstruction by NeRF. Fillioux~\textit{et al.}~\cite{fillioux2023spatio} evaluate SAM's performance on processing patient-derived organoids microscopy frames. MediViSTA-SAM~\cite{kim2023medivista} is the first study on adapting SAM to video segmentation. SuPerPM~\cite{lin2023superpm} is a large deformation-robust surgical perception framework, which utilizes SAM to segment tissue regions from the background.

\subsubsection{Domain Adaptation}
Recently, researchers utilized SAM to enhance the generalization ability of the model on target domain, especially in situations where the quality and quantity of data in the target domain are less than ideal. Bonani \textit{et al.}\cite{bonani2023learning} utilized SAM to provide a regularization signal for real data and introduced an invariance-variance loss structure. This structure is defined for self-supervised learning on unlabeled target domain data, facilitating the robustness of domain adaptation ability for semantic segmentation networks. Yao \textit{et al.}\cite{yao2023sam} proposed SAM-DA, a SAM-powered domain adaptation framework designed for real-time nighttime UAV tracking. They introduced an innovative SAM-driven method to expand target domain training samples, which generates a substantial quantity of high-quality training samples for the target domain from each nighttime image, enabling one-to-many sample generation. This approach significantly augments both the quantity and quality of target domain training samples, thereby providing improved data support for domain adaptation.

\subsubsection{Tool Software} 
Hsieh \textit{et al.}\cite{hsieh2023tool} explored the possibility of leveraging tool documentation, as opposed to demonstrations, for instructing large language models (LLMs) on the utilization of new tools. The article~\cite{hsieh2023tool} demonstrated that the use of tool documentation empowered LLMs to employ SAM in a zero-shot manner, eliminating the need for training or finetuning. Of equal significance, the article showcased the potential of employing tool documentation to enable novel applications. One such illustration involved the amalgamation of GroundingDino\cite{liu2023grounding} and SAM, resulting in the creation of Grounded-SAM\cite{groundSAM2023iccvdemo}—a model proficient in generating text grounded in visual content, showcasing its capabilities to derive meaningful textual information from images.

\subsubsection{More Directions} 
Several studies have applied SAM in various applications, spanning optical flow estimation\cite{zhou2023samflow}, robotics\cite{yang2023pave, kannan2023learning, benjdira2023rosgpt_vision}, reinforcement learning (RL) for video games\cite{schiller2023virtual}, and semantic communication\cite{raha2023generative}.

To address the challenge of ``fragmentation" in optical flow estimation, Zhou \textit{et al.}\cite{zhou2023samflow} employed SAM as an image encoder, providing optical flow estimation with richer and higher-level contextual features. This strategy mitigates the model's tendency to focus exclusively on local and low-level cues. In the work by Yang \textit{et al.}\cite{yang2023pave}, SAM was used to generate segmentation masks for objects, providing the model with rich semantic, geometric, and shape priors. This, in turn, assists robots in perceiving object poses and determining grasp points. Similar ideas are also evident in \cite{kannan2023learning, benjdira2023rosgpt_vision}. In \cite{schiller2023virtual}, the authors enhanced the original pixel input using SAM, aiming to improve the performance of RL agents in Atari video games. Despite the observed improvement in the game-playing performance of the RL agent, finding a suitable balance between performance enhancement and computational cost remains an ongoing exploration. Additionally, Raha \textit{et al.}\cite{raha2023generative} proposed a novel semantic communication framework based on SAM, efficiently transmitting sequential images or videos while preserving the original content unchanged.

\begin{figure*}[t]
\centering
\includegraphics[width =1.0\linewidth]{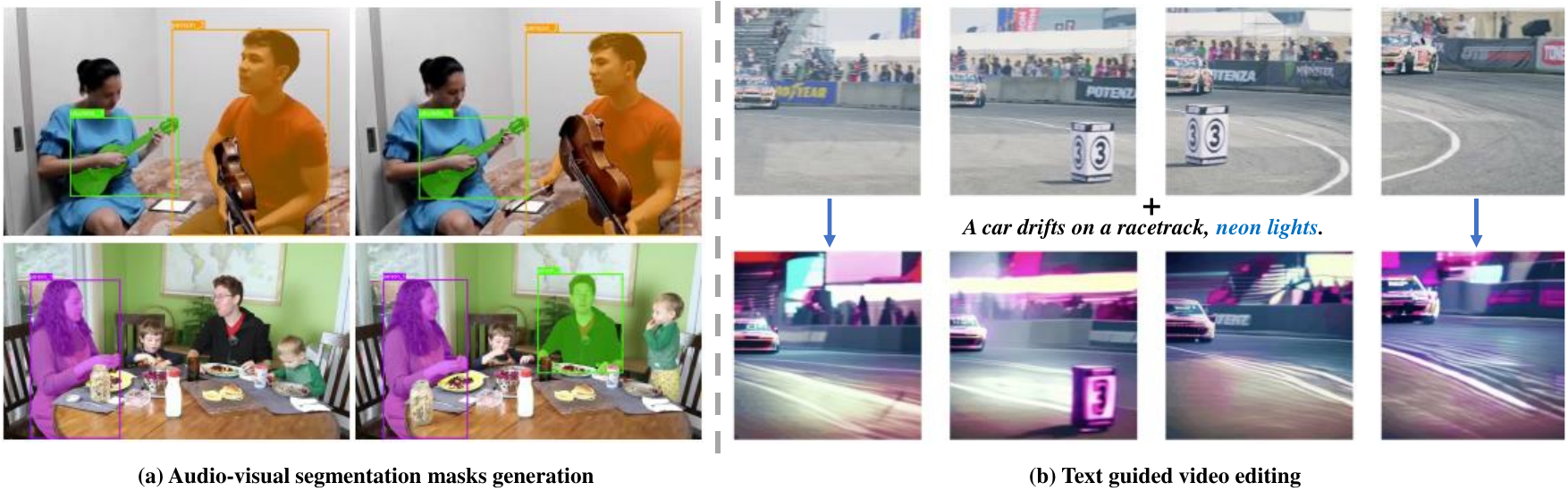}

\caption{Examples of video masks generation results with (a) AVISeg~\cite{guo2023audio} and video editing results with (b) 2SVE~\cite{wu2023cvpr}.}
\label{fig:video_generation_and_editing}

\end{figure*}

%------------------------------------------------------------------------------
\section{Video Generation with SAM}
\label{sec:class_2}

In this section, we divide video generation with SAM into four groups and provide detailed reviews for each: video synthesis (\eg, dance generation)~\cite{qin2023dancing,wang2023disco}, video super-resolution (\ie, generating more detailed and visually appealing videos from low-resolution versions)~\cite{Lu2023CanSB}, 3D reconstruction (\eg, reconstruction and segmentation of 3D objects and producing point-level semantic labels for 3D point cloud)~\cite{yang2023sam3d,xu2023one}, and video dataset annotation generation (\eg, bounding boxes and masks generation)~\cite{balaban2023propagating,videoTextSpottingSA,guo2023audio,delatolas2023learning,kiefer2023stable}. The taxonomy details of video generation with SAM is illustrated in Fig.~\ref{fig:sam_based_methods}.

\subsection{Video Synthesis}
SAM has been recently employed in two works focusing on dance video synthesis \cite{qin2023dancing, wang2023disco}.
In the Dancing Avatar project, SAM is employed to generate contextually appropriate background images for human motion videos, following textual specifications and using image inpainting techniques. This method leads to the creation of distinct pose-guided images for various poses, resulting in the generation of human area masks. In the research presented in \cite{qin2023dancing}, SAM ensures a consistent background throughout the human motion image sequence, effectively separating the human foreground from the background \cite{wang2023disco}. SAM's outstanding performance in these projects, along with its contributions to other modules, has played a pivotal role in achieving an impressive result.

\subsection{Video Super-Resolution} 
The main challenge in video super-resolution (VSR) lies in handling large motions in input frames, making it challenging to accurately aggregate information from multiple frames. However, according to literature \cite{Lu2023CanSB}, existing methods overlook valuable semantic information that could significantly enhance results, and flow-based approaches heavily depend on the accuracy of flow estimates, which may be imprecise for two low-resolution frames.

In~\cite{Lu2023CanSB}, a robust and semantic-aware prior for improved VSR was investigated by leveraging the SAM. To incorporate the SAM-based prior, the study proposed the SAM-guidEd refinEment Module (SEEM), a simple yet effective module enhancing both alignment and fusion procedures through the utilization of semantic information. This lightweight plug-in module is designed not only to leverage attention mechanisms for generating semantic-aware features but also to be easily integrated into existing methods. Specifically, SEEM was applied to two representative methods, EDVR and BasicVSR, resulting in consistently improved performance with minimal implementation effort across three widely used VSR datasets: REDS, Vid4, and Vimeo-90K~\cite{xue2019video}. Importantly, SEEM was found to enhance existing methods efficiently, providing increased flexibility in adjusting the balance between performance and the number of training parameters.

\subsection{3D Reconstruction} 
Recent research has explored leveraging SAM's robust generalization and transfer capabilities to extend its application from 2D image segmentation to tasks related to 3D reconstruction and segmentation. These efforts bring forth a fresh perspective and methodology for understanding and reconstructing 3D scenes. Based on the approaches of converting the 2D segmentation masks into 3D masks, existing methodologies can be broadly classified into two categories: tracking-based approaches \cite{xu2023one} and projection-based approaches \cite{yang2023sam3d, dong2023leveraging}.

\myPara{Tracking-based Approaches.} Xu \textit{et al.}\cite{xu2023one} introduced OSTRA, an open-source one stop 3D target reconstruction and multilevel segmentation framework. Within this framework, SAM is employed to segment the first frame of a video. Subsequently, the authors use VOT algorithms to generate continuous masks for video frames. In this process, SAM is also used to complementarily correct tracking errors. Finally, 3D reconstruction methods are applied to reconstruct labeled 3D objects or multiple components. Experimental results demonstrate that OSTRA can support common 3D object models, including point clouds, meshes, and voxels. Meanwhile, in complex scenes with intricate structures and occlusions, OSTRA outperforms manual segmentation.

\myPara{Projection-based Approaches.} The procedural framework of projection-based approaches involves three key steps: firstly, applying SAM to segment RGB images of the 3D scene; secondly, projecting 2D segmentation masks into 3D space; and finally, generating 3D semantic pseudo labels. Yang \textit{et al.}\cite{yang2023sam3d} introduced SAM3D, a 3D point cloud segmentation framework based on SAM. In the step of generating 3D semantic pseudo labels, SAM3D incorporates a bottom-up merging method, iteratively combining 3D masks from different frames, thereby consolidating the masks from two adjacent frames into a singular representation. Diverging from the aforementioned merging method\cite{yang2023sam3d}, Dong \textit{et al.}\cite{dong2023leveraging} proposed a cumulative semantic fusion (CSF) method. This approach integrates 3D segmentation results from various perspectives through a voting mechanism to generate 3D pseudo-semantic labels. Furthermore, the CSF framework addresses diverse scenarios with three segmentation strategies: a CLIP-based LSeg\cite{Li2022Lseg} strategy, a SAM with Grounding-DINO\cite{liu2023grounding} strategy, and a SAM with 2D sparse point annotations strategy.

\subsection{Video Dataset Annotation Generation} 
Due to the high cost associated with annotating videos in specific domains, many datasets lack effective labels, particularly at the pixel level. Some studies have harnessed the advantage of SAM to design systems for automatically annotating video data, providing adaptability to various scenes and objects. Balaban \textit{et al.}\cite{balaban2023propagating} proposed a semantic label propagation (SLP) system that integrates SAM and structure from motion (SfM) for automatic video data annotation. SAMText\cite{videoTextSpottingSA} is a scalable mask annotation pipeline capable of rapidly generating fine mask annotations for scene text images or video frames at scale. EVA-VOS\cite{delatolas2023learning} employs an intelligent agent to iteratively predict frames to annotate and the annotation types to use, establishing a human-in-the-loop annotation framework for video object segmentation. Experimental results indicate that EVA-VOS's annotation method achieves segmentation quality close to human consistency and is 3.5 times faster than traditional annotation methods.

Some studies have also leveraged SAM to introduce innovative datasets\cite{kiefer2023stable}. SAMText-9M\cite{videoTextSpottingSA} is a novel large-scale video text detection dataset, comprising over 2,400 video clips and more than 9 million segmentation masks. In contrast to data annotated with text position using quadrilateral bounding boxes, SAMText-9M utilizes detailed segmentation masks for text annotation. In the domain of audio-visual instance segmentation, Guo \textit{et al.}\cite{guo2023audio} constructed the first audio-visual instance segmentation dataset, AVISeg (see Fig.~\ref{fig:video_generation_and_editing}(a)), which includes 1,258 videos, 78,665 frames, 281 objects, 15,355 masks, and 26 categories. Additionally, AVISeg can be applied to various multi-modal video understanding tasks such as video editing, virtual reality, and robot navigation.

\begin{comment}
\begin{table}[t]
  \centering
  %\captionsetup{font={small}}
  \caption{Statistics of some video datasets constructed using SAM. (1K = 1000, 1M = 1000K).}
  %\footnotesize
  \setlength{\tabcolsep}{0.15mm}{
  \begin{tabular}{ccccc} 
    \Xhline{1.2pt}  % 表格横线加粗
    Datasets& Task field&  Year &  Clips  & Masks  \\ \hline
    SAMText-9M\cite{videoTextSpottingSA}  &Video Text Spotting &2023&  2400 &  9M \\ 
    AVISeg\cite{guo2023audio}  &Audio-Visual Instance Segmentation& 2023& 1258  &  15355 \\
    DSEC-MOS& & & & \\
   \Xhline{1.2pt}  % 表格横线加粗
  \end{tabular}
  }
  \label{tab:dataset}
\end{table}
\end{comment}

%------------------------------------------------------------------------------
\section{Video Editing with SAM}
\label{sec:class_3}

In this section, we detail the video editing algorithms using SAM that are divided into three groups: generic video editing, text guided video editing, and object removing. The taxonomy relations of video editing with SAM are illustrated in Fig.~\ref{fig:sam_based_methods}. 

\myPara{Generic Video Editing.} Make-A-Protagonist~\cite{zhao2023make} proposes a framework for generic video editing with both visual and textual clues. It leverages multiple pre-trained experts to process source video and target visual/textual clues. Then, all the information is put into the proposed visual-textual-based video generation model with mask-guided denoising sampling to generate the desired output. SAM plays a key role in segmenting the protagonist based on the text description and masking out the background in the reference image.

\myPara{Text Guided Video Editing.} Wu \textit{et al.}~\cite{wu2023cvpr} introduces a new dataset (TGVE) that contains 76 videos with 4 prompts each for text-guided video editing (see Fig.~\ref{fig:video_generation_and_editing}(b)). Based on the TGVE dataset, the competition workshop was held at CVPR 2023. The winning method Two-Stage Video Editing (2SVE) incorporates many pre-trained models such as SAM, OpenCLIP~\cite{ilharco2021openclip} and ControlNet~\cite{zhang2023adding}. 
The target segment process is based on the SAM and OpenCLIP models. It works as follows: SAM predicts the masks of the input frame automatically. Then, OpenCLIP converts the masks into embeddings and calculates similarity with the text embeddings to select the target mask for the next steps. The 2SVE method consists of two stages, as its name indicates. The first stage uses ControlNet to edit the foreground, background, and structure of the input video. The second stage uses a diffusion model trained on the MSVD~\cite{chen-dolan-2011-collecting} dataset to edit the style and appearance of the output video. ControlNet and diffusion model take the target masks from the target segment process as guidance in both stages.

\myPara{Object Removing.} Researchers also leverage the strong prompt segmentation ability of SAM to remove objects in 3D scenes. OR-NeRF~\cite{yin2023or} proposes a novel object-removing pipeline using either points or text prompts on a single view and ensuring multiview consistency and plausible completion after deletion. OR-NeRF consists of two stages: multiview segmentation and scene object removal. For the input in the first stage, the model either uses point prompts directly or converts the input text information into point prompts via Ground-SAM. The model uses SAM to predict the mask of images from all viewing angles based on prompts and uses LaMa to obtain color and depth priors. Then, using NeRF to reconstruct the scene after removal. Based on their one-step multiview segmentation method, which leverages SAM's strong power, it achieves better removal quality and requires less time than previous methods.

\begin{table*}[t]
  \centering
  %\captionsetup{font={small}}
  \caption{Performance evaluation of different VOS tasks, including video semantic segmentation (DAVIS 2017 val~\cite{pont20172017} and DAVIS 2016 val~\cite{perazzi2016benchmark} datasets), zero-shot open-world video instance segmentation (UVO~\cite{wang2021unidentified}), and zero-shot video instance segmentation (HQ-YTVIS~\cite{ke2022video}). The best results are marked in \textbf{bold}.}
 
  %\footnotesize
  \setlength{\tabcolsep}{0.5mm}{
  \begin{tabular}{ccccccccccc} 
    \Xhline{1.2pt}  % 表格横线加粗
    \rowcolor[gray]{0.9} \multicolumn{11}{c}{\textbf{Video Semantic Segmentation}} \\ 
    \hline
     \multirow{2}{*}{Method} & \multirow{2}{*}{Venue} & \multirow{2}{*}{SAM-Based} & \multirow{2}{*}{Training Data}  & \multirow{2}{*}{Prompt}  & \multicolumn{3}{c}{DAVIS 2017 val}  & \multicolumn{3}{c}{DAVIS 2016 val} \\
    & & & & &  $J\&F$ & $J$ & $F$ &  $J\&F$ & $J$ & $F$ \\

    \hline
    %AGAME~\cite{johnander2019generative}  &  CVPR 2019 &  \ding{55} & Video~\cite{johnander2019generative}  &  - &  70.0 & 67.2   &  72.7 &- & - &  -\\
    
    %AGSS~\cite{lin2019agss}  &  ICCV 2019 &  \ding{55} & Video~\cite{}   & - & 67.4 & 64.9 & 69.9 &- & - &  -\\

    %AFB-URR~\cite{liang2020video}  &  NeurIPS 2020 &  \ding{55} & Video~\cite{}   & - & 74.6 & 73.0 & 76.1 &- & - &  -\\

    %SWEM~\cite{lin2022swem}  &  CVPR 2022 &  \ding{55} & Video~\cite{}   & - & 84.3 & 81.2 & 87.4 &  91.3 & 89.9  & 92.6  \\

    %AOT~\cite{yang2021associating}  &  NeurIPS 2022 &  \ding{55} & Video~\cite{xu2018youtube,pont20172017}   & - & 85.4 & 82.4 & 88.4 & 92.0 & 90.7  & 93.3  \\
    
    XMem~\cite{cheng2022xmem}  &  ECCV 2022 &  \ding{55} & Video~\cite{pont20172017,xu2018youtube}   & - & \textbf{87.7} & \textbf{84.0} & \textbf{91.4} & \textbf{92.0} & \textbf{90.7}  & \textbf{93.2} \\
    
    %DEVA~\cite{cheng2023tracking}  &  ICCV 2023 &  \ding{55} & Video~\cite{}   & - & 73.4 & - & - & 88.9 & -  & -  \\

    %Painter~\cite{wang2023images}  &  CVPR 2023 &  \ding{55} & Image~\cite{}   & - & 34.6 & 28.5 & 40.8 & 70.3 & 69.6 & 70.9 \\

    %SegGPT~\cite{wang2023seggpt}  &  ICCV 2023 &  \ding{55} & Image~\cite{wang2023seggpt}   & Image+Mask & 75.6 & 72.5  & 78.6  & 70.9 & 83.6  & 83.8  \\

    PerSAM~\cite{zhang2023personalize}  &  arXiv 2023 &  \ding{52} & Training-free    & Mask & 60.3 & 56.6  & 63.9  & - & -  & -  \\

    PerSAM-F~\cite{zhang2023personalize}  &  arXiv 2023 &  \ding{52} & Image~\cite{ruiz2023dreambooth}    & Mask & 71.9  & 69.0  & 74.8  & - & -  & -  \\
      
    UVOSAM~\cite{zhang2023uvosam}  &  arXiv 2023 &  \ding{52} & Video~\cite{pont20172017} & Trajectories & 78.9 & 75.5  & 82.0  & - & -  & -  \\

    Matcher~\cite{liu2023matcher}  &  arXiv 2023 &  \ding{52} & Training-free & Point/Center/Box & 79.5 & 76.5  & 82.6  & 86.1 & 85.2  & 85.2  \\

    GT Box+SAM~\cite{ICCV2023SAM}  &  ICCV 2023 &  \ding{52} & Training-free & Box & 87.3 & 83.5  & 91.0  & - & -  & -  \\

    %%%%%%%%%%%%%%%%%%%%%%%%%%%%%%%%%%%%%%%%%%%%%%%%%%%%%%%%%%%%%%%%%
    \Xhline{1.2pt}  % 表格横线加粗
    
    \rowcolor[gray]{0.9} \multicolumn{11}{c}{\textbf{Zero-shot Open-world Video Instance Segmentation}} \\ 
    %\rowcolor[gray]{0.9} \multicolumn{11}{c}{\textbf{Video Instance Segmentation}} \\ 
    \hline
    \multirow{2}{*}{Method} & \multirow{2}{*}{Venue} & \multirow{2}{*}{SAM-Based} & \multirow{2}{*}{Training Data}  & \multirow{2}{*}{Prompt}  & \multicolumn{6}{c}{UVO} \\
      & & &  & & ${\rm AP}_{B}^{\rm strict}$ & ${\rm AP}_{B75}^{\rm strict}$ & ${\rm AP}_{B50}^{\rm strict}$ & ${\rm AP}_{B}$ & ${\rm AP}_{B75}$ & ${\rm AP}_{B50}$\\

    \hline
    DTM~\cite{du2021uvo}  &  arXiv 2023 &  \ding{55} & Image~\cite{lin2015microsoft,krasin2017openimages,everingham2010pascal}+Video~\cite{wang2021unidentified}   &  -  & - & - & - & \textbf{27.6} & \textbf{29.2}  & \textbf{40.6}  \\
        
    SAM~\cite{ICCV2023SAM}  &  ICCV 2023 &  \ding{52} & Image~\cite{ICCV2023SAM}   &  Box  & 8.6 & 3.7  & 25.6 & 17.3 & 14.4  & 37.7  \\

    HQ-SAM~\cite{ke2023segment}  &  arXiv 2023 &  \ding{52} & Image~\cite{ke2023segment}    &  Box  & \textbf{9.9} &  \textbf{5.0}  & \textbf{28.2}  & 18.5 &16.3 &38.6\\

    \Xhline{1.2pt}  % 表格横线加粗
    \rowcolor[gray]{0.9} \multicolumn{11}{c}{\textbf{Zero-shot Video Instance Segmentation}} \\ 
    \hline 
    %\hline 
    
    \multirow{2}{*}{Method} & \multirow{2}{*}{Venue} & \multirow{2}{*}{SAM-Based} & \multirow{2}{*}{Training Data}  & \multirow{2}{*}{Prompt}  & \multicolumn{6}{c}{HQ-YTVIS} \\
      & & &  & & ${\rm AP}^{B}$ & ${\rm AP}_{75}^{ B}$ & ${\rm AP}_{50}^{ B}$ & ${\rm AP}^{M}$ & ${\rm AP}_{75}^{ M}$ & ${\rm AP}_{50}^{ M}$\\
      
    \hline
    %SeqFormer~\cite{wu2021seqformer}  &  ECCV 2022 &  \ding{55} & Image~\cite{lin2014microsoft}+Video~\cite{yang2019video}   &  -  & 43.3 & 41.5 & - & 63.7 & 69.7 & -  \\

    VMT~\cite{ke2022video}  &  ECCV 2022 &  \ding{55} & Video~\cite{ke2022video}   &  -  & \textbf{44.8} & \textbf{43.4}  & - & \textbf{64.8}  & 70.1 & -   \\
      
    SAM~\cite{ICCV2023SAM}  &  ICCV 2023 &  \ding{52} & Image~\cite{ICCV2023SAM}   &  Box  & 30.2 & 19.1 &72.9& 60.7& 68.1& 90.5  \\

    HQ-SAM~\cite{ke2023segment}  &  arXiv 2023 &  \ding{52} & Image~\cite{ke2023segment}    &  Box  & 34.0& 24.3& \textbf{79.5} & 63.6 & \textbf{70.5} & \textbf{91.1}\\

   \Xhline{1.2pt}  % 表格横线加粗

  \end{tabular}
  }
  \label{tab:videoobjectsegmentation}

\end{table*}

%------------------------------------------------------------------------------
\section{Performance Evaluation}
\label{sec:PerformanceEvaluation}

In this section, we introduce the benchmark datasets, evaluation metrics, and comparative results of current SOTA and SAM-based methods across different video tasks.

\subsection{Evaluation of Video Object Segmentation Approaches}
\myPara{Datasets.} DAVIS 2016~\cite{perazzi2016benchmark} and DAVIS 2017~\cite{pont20172017} are two widely used datasets for VOS. DAVIS 2016 contains 50 videos with a total of 50 annotated object instances. The dataset is split into 30 videos for training and 20 videos for validation. DAVIS 2017 is an extension of~\cite{perazzi2016benchmark}, with toltaling 150 videos and 376 annotated object instances. The test and validation sets of this dataset both contain 30 videos. Unidentifed Video Objects (UVO)~\cite{wang2021unidentified} is a large-scale dataset for open-world object segmentation in videos. HQ-YTVIS~\cite{ke2022video} is a video dataset for high-quality video instance segmentation, including 1,678 videos for the training set, 280 videos for the validation set, and 280 videos for the test set.

\myPara{Evaluation Metrics.} In the evaluation of the video semantic segmentation task, the commonly used metrics are region ($\mathcal{J}$) and contour ($\mathcal{F}$) measures proposed in DAVIS 2016~\cite{perazzi2016benchmark},  and $\mathcal{J\&F}$ metric proposed in DAVIS 2017~\cite{pont20172017} by calculating the mean of region similarity and contour accuracy over all object instances. The boundary AP$_{B}$ and stricter AP$_{B}^{\rm strict}$~\cite{ke2023segment} are adopted to assess the mask quality for the UVO dataset. For the HQ-YTVIS dataset~\cite{ke2022video}, the evaluation metrics include the standard tube mask AP$^{M}$ and Tube-Boundary AP$^{B}$.

\myPara{Results Comparison.} For VOS, we mainly conduct comparison on video semantic segmentation and video instance segmentation as there are numbers of methods for benchmarking.
In Tab.~\ref{tab:videoobjectsegmentation}, we present VOS performance of current SOTA and SAM-based methods on DAVIS 2017 val~\cite{pont20172017}, DAVIS 2016 val~\cite{perazzi2016benchmark}, UVO~\cite{wang2021unidentified}, and HQ-YTVIS~\cite{ke2022video} datasets.

The main observations are as follows: \textbf{1)} The SAM-based methods have significant performance gaps compared to the current SOTA methods designed for specific video segmentation tasks. This is because many SAM-based methods directly utilize SAM pre-trained on image data to enhance the ability of object segmentation, ignoring the importance of end-to-end training and finetuning for inherently complex video tasks. For instance, on DAVIS 2017 val~\cite{pont20172017}, the best SAM-based video semantic segmentation method (Matcher~\cite{liu2023matcher}) is training free, achieving a $\mathcal{J \& F}$ score of 79.5\%, while Xmen~\cite{cheng2022xmem} obtains a $\mathcal{J \& F}$ score of 87.7\%. Xmen is a long-term VOS method with multiple feature memory stores. This highlights the importance of memory mechanism for ensuring temporal consistency and coherence of the predicted masks across long videos. \textbf{2)} The high-quality of prompt is crucial for the zero-shot generalization ability of SAM. One example is that ``GT Box+SAM''~\cite{ICCV2023SAM} uses reliable ground-truth bounding boxes as prompts and achieves the second best results on DAVIS 2017 val. \textbf{3)} The models (\eg, XMem~\cite{cheng2022xmem} and VMT~\cite{ke2022video}) trained on video data exhibit significant advantages over the models trained solely on image data, and the utilization of multi-modal data (\eg, video and image)~\cite{du2021uvo} often leads to improved performance.

\begin{table}[t]
  \centering
  %\captionsetup{font={small}}
  \caption{Performance evaluation on six VOT datasets (VOTS2023~\cite{kristan2023first}, TREK-150~\cite{dunnhofer2023visual}, NUT-L~\cite{yao2023sam}, DAVIS 2017 test~\cite{pont20172017}, DAVIS 2016 val~\cite{perazzi2016benchmark}, and YouTube-VOS 2018 val~\cite{xu2018youtube}). We summarize current SOTA and SAM-based methods. The best results are marked in \textbf{bold}.}

  %\footnotesize
  \setlength{\tabcolsep}{0.2mm}{
  \begin{tabular}{ccccccccc} 

    \Xhline{1.2pt}  % 表格横线加粗
    \rowcolor[gray]{0.9} \multicolumn{6}{c}{\textbf{Video Object Tracking}} \\ 
    \hline

    \multirow{2}{*}{Method} & \multirow{2}{*}{SAM-Based} & \multirow{2}{*}{Initialization}  & \multicolumn{3}{c}{VOTS2023} \\
      & & &  $AUC$ & $A$ & $R$  \\
      
    \hline
    DMAOT~\cite{kristan2023first}  &  \ding{55} &  Mask  & \textbf{63.6} &75.1 & \textbf{79.5} \\

    %VOMS~\cite{yang2022decoupling}+SAM-H  &  \ding{52}  &  Mask  & 61.0  & 75.1 &  75.7 \\

    HQTrack~\cite{zhu2023tracking}  &  \ding{52}  &  Mask  & 61.5  & \textbf{75.2} &  76.6 \\
    \hline
    \hline

    \multirow{2}{*}{Method} & \multirow{2}{*}{SAM-Based} & \multirow{2}{*}{Initialization}  & \multicolumn{3}{c}{TREK-150 test set} \\
      & & &  $MSE$ & $OPE$ & $HOI$  \\
      
    \hline
    LTMU-H-IJCV~\cite{dai2020high}  &  \ding{55} &  Box  & 54.3 &   50.5  & 65.7  \\

    MSDeAOT~\cite{xu2023zju}  &  \ding{52}  &  Box  & \textbf{73.4}  &  \textbf{75.5} &  \textbf{77.1} \\

    \hline
    \hline

    \multirow{2}{*}{Method} & \multirow{2}{*}{SAM-Based} & \multirow{2}{*}{Initialization}  & \multicolumn{3}{c}{NUT-L} \\
      & & &  $AUC$ & $nPre$ & $Pre$  \\
      
    \hline
    UDAT~\cite{ye2022unsupervised}  &  \ding{55} &  Box  & 37.7 &  43.4 & 49.8 \\

    SAM-DA~\cite{yao2023sam}  &  \ding{52}  &  Box  & \textbf{43.0}  & \textbf{49.2} &  \textbf{56.4} \\

    \hline
    \hline
    
    \multirow{2}{*}{Method} & \multirow{2}{*}{SAM-Based} & \multirow{2}{*}{Initialization}  & \multicolumn{3}{c}{DAVIS 2017 test} \\
      & & &  $J\&F$ & $J$ & $F$  \\
      \hline

    %MiVOS~\cite{cheng2021modular}  &  \ding{55} &  Scribble  & 78.6 & 74.9  & 82.2 \\
        
    %XMem~\cite{cheng2022xmem}  &  \ding{55} &  Mask  & 81.2 & 77.6 & 84.7 \\

    SwinB-DeAOT-L~\cite{yang2022decoupling}  &  \ding{55} &  Mask  & \textbf{82.8} & \textbf{78.9} & \textbf{86.7} \\

    TAM~\cite{yang2023track}  &  \ding{52}  &  Click  & 73.1  & 69.8 &  76.4 \\

    SAM-Track~\cite{cheng2023segment}  &  \ding{52}  &  Click  & 79.2  & 75.3 &  83.1 \\
    \hline
    \hline
    
    \multirow{2}{*}{Method} & \multirow{2}{*}{SAM-Based} & \multirow{2}{*}{Initialization}  & \multicolumn{3}{c}{DAVIS 2016 val} \\
      & & &  $J\&F$ & $J$ & $F$  \\
      
    \hline
    %MiVOS~\cite{cheng2021modular}  &  \ding{55} &  Scribble  & 91.0 & 89.6 & 92.4\\
        
    %XMem~\cite{cheng2022xmem}  &  \ding{55} &  Mask  & 92.0 & 90.7 & 93.2 \\

    SwinB-DeAOT-L~\cite{yang2022decoupling}  &  \ding{55} &  Mask  & \textbf{92.9} & \textbf{91.1} & \textbf{94.7} \\

    TAM~\cite{yang2023track}  &  \ding{52}  &  Click  & 88.4  & 87.5 &  89.4 \\

    SAM-Track~\cite{cheng2023segment}  &  \ding{52}  &  Click  & 92.0  & 90.3 &  93.6 \\

    \hline
    \hline
    
    \multirow{2}{*}{Method} & \multirow{2}{*}{SAM-Based} & \multirow{2}{*}{Initialization}  & \multicolumn{3}{c}{YouTube-VOS 2018 val} \\
    
      & & & $J\&F$ & $J$ & $F$  \\
      \hline

   STCN~\cite{cheng2021rethinking}  &  \ding{55} &  Mask  & \textbf{83.0} & \textbf{81.9} & \textbf{86.5} \\
   
    %Detic~\cite{zhou2022detecting}+STCN~\cite{cheng2021rethinking}  &  \ding{55} &  Detected Mask  & 58.8 & 68.1 & 71.7 \\
        
    OVTracktor~\cite{chu2023zero}  &  \ding{52} &  Detected Mask  & 62.2  & 65.9  & 69.4   \\

    PerSAM-F~\cite{zhang2023personalize}  &  \ding{52} &  Mask  & 71.9  & 69.0  & 74.8   \\

    SAM-PT~\cite{rajivc2023segment}  &  \ding{52} &  Query Points  & 76.3  & 73.6 & 78.9    \\

    SAM-PT-reinit~\cite{rajivc2023segment}  &  \ding{52} &  Refining Points  & 76.6   & 74.4  & 78.9  \\

    HQ-SAM-PT~\cite{ke2023segment}  &  \ding{52} &  Query Points  & 77.2   & 74.7   & 79.8   \\

    HQ-SAM-PT-reinit~\cite{ke2023segment}  &  \ding{52} &   Refining Points  & 77.0   & 77.0   & 79.2  \\

   \Xhline{1.2pt}  % 表格横线加粗

  \end{tabular}
  }
  \label{tab:videoobjecttracking}

\end{table}

\begin{table}[t]
  \centering
  %\captionsetup{font={small}}
  \caption{Performance evaluation of other video understanding tasks, including deepfake detection, video shadow detection, miscellaneous (\ie, audio-visual segmentation and RVOS), and domain specific (\ie, optical flow estimation and 3D point cloud segmentation) methods.}

  %\footnotesize
  \setlength{\tabcolsep}{0.0001mm}{
  \begin{tabular}{cccccccc}

    \Xhline{1.2pt}  % 表格横线加粗
    \rowcolor[gray]{0.9} \multicolumn{8}{c}{\textbf{Deepfake Detection}} \\ 
    \hline 

    \multirow{2}{*}{Method} & \multirow{2}{*}{SAM-Based}  & \multicolumn{5}{c}{FaceForensics++} \\
    
     &  & DF  & F2F & FS & NT & \multicolumn{2}{c}{Average} \\

    \hline
    Locate~\cite{kong2022detect}  &  \ding{55}  & 97.25  & 94.46 & 97.13 & 84.63 &  \multicolumn{2}{c}{93.36 }\\

    %SAM~\cite{ICCV2023SAM}  &   \ding{52} & 89.32 & 84.56 & 91.19 & 80.01 & \multicolumn{2}{c}{86.27}\\

    DADF~\cite{lai2023detect}  &   \ding{52} & \textbf{99.02} & \textbf{98.92} & \textbf{98.23} & \textbf{87.61} & \multicolumn{2}{c}{\textbf{95.94}}\\
    
    \Xhline{1.2pt}  % 表格横线加粗
    \rowcolor[gray]{0.9} \multicolumn{8}{c}{\textbf{Video Shadow Detection}} \\ 
    \hline 
    \multirow{2}{*}{Method} & \multirow{2}{*}{SAM-Based}  & \multicolumn{5}{c}{ViSha} \\
    
      & & MAE & F$_\beta$ & IoU & SBER & \multicolumn{2}{c}{NBER} \\
      
    \hline
    Liu \emph{et al.}~\cite{liu2023scotch}  &  \ding{55}  & 0.029 & 0.793 & 0.640 &  16.26 & \multicolumn{2}{c}{1.44}\\

    ShadowSAM~\cite{wang2023detect}  &   \ding{52} & \textbf{0.024} &  \textbf{0.813} & \textbf{0.661} & \textbf{25.21} &  \multicolumn{2}{c}{\textbf{1.13}  } \\

    \Xhline{1.2pt}  % 表格横线加粗
    \rowcolor[gray]{0.9} \multicolumn{8}{c}{\textbf{Audio-Visual Segmentation}} \\ 
    \hline 
        \multirow{2}{*}{Method} & \multirow{2}{*}{SAM-Based}  & \multicolumn{3}{c}{AVSBench-V1S}   & \multicolumn{3}{c}{AVSBench-V1M} \\
    
      & &  mIoU & \multicolumn{2}{c}{F-score}  & mIoU &  \multicolumn{2}{c}{F-score}\\
      
    \hline
    AUSS~\cite{ling2023hear}  &  \ding{55} & \textbf{89.4} &  \multicolumn{2}{c}{\textbf{94.2}}  &63.5 & \multicolumn{2}{c}{75.2} \\

    %SAM~\cite{ICCV2023SAM}  &   \ding{52} &  29.7 & \multicolumn{2}{c}{42.1} & - & \multicolumn{2}{c}{-} \\

    AV-SAM~\cite{mo2023av}  &   \ding{52} & 40.8 & \multicolumn{2}{c}{56.6} & - & \multicolumn{2}{c}{-}  \\

    CMSF~\cite{bhosale2023leveraging}  &   \ding{52} & 58.0& \multicolumn{2}{c}{67.0}  &34.0& \multicolumn{2}{c}{44.0} \\
    
    GAVS~\cite{wang2023prompting}  &   \ding{52} & 80.1 & \multicolumn{2}{c}{90.2} & \textbf{63.7} & \multicolumn{2}{c}{\textbf{77.4}} \\

    \Xhline{1.2pt}  % 表格横线加粗
    \rowcolor[gray]{0.9} \multicolumn{8}{c}{\textbf{Referring Video Object Segmentation}} \\ 
    \hline 
        \multirow{2}{*}{Method} & \multirow{2}{*}{SAM-Based}  & \multicolumn{3}{c}{Ref-DAVIS17} & \multicolumn{3}{c}{Ref-YouTube-VOS} \\
    
      & &  $J\&F$ & $J$ & $F$ &  $J\&F$ & $J$ & $F$ \\
      
    \hline
    ReferFormer~\cite{wu2022language}  &  \ding{55}  & 61.1 & 58.1 & 64.1 & \textbf{64.9} & \textbf{62.8} & \textbf{67.0}   \\

    RefSAM~\cite{li2023refsam}  &   \ding{52} & \textbf{66.1} & \textbf{62.9}  & \textbf{69.3} & 55.1  &   53.9   &   56.3\\

    \Xhline{1.2pt}  % 表格横线加粗
    \rowcolor[gray]{0.9} \multicolumn{8}{c}{\textbf{Medical Videos}} \\ 
    \hline 
    \multirow{2}{*}{Method} & \multirow{2}{*}{SAM-Based}  & \multicolumn{3}{c}{EndoVis2017} & \multicolumn{3}{c}{EndoVis2018} \\
    
      & & cIoU & IoU & mcIoU &cIoU & IoU & mcIoU \\
      
    \hline
    MATIS Full~\cite{ayobi2023matis}  &  \ding{55}  & 71.36 & 66.28 & 41.09 & \textbf{84.26} & 79.12 & 54.04\\

    SurgicalSAM~\cite{yue2023surgicalsam}  &   \ding{52} & 69.94 &  \textbf{69.94} & \textbf{67.03} & 80.33 &\textbf{80.33}& \textbf{58.87} \\

     %SAM 1 Point~\cite{wang2023sam}  &   \ding{52} & 55.96 &-& - & 54.30 &-& -\\
    
    SAM Box~\cite{wang2023sam}  &   \ding{52} & \textbf{88.20} &- & - & 81.09 &- &- \\

    \hline
    \hline
    \multirow{2}{*}{Method} & \multirow{2}{*}{SAM-Based}  & \multicolumn{3}{c}{CAMUS} & \multicolumn{3}{c}{In-house Data} \\
    
      & & Dice & dH & dA & Dice & dH & dA \\
      
    \hline
    SwinUNETR~\cite{hatamizadeh2021swin}  &  \ding{55}  & 94.0 & 5.02 &  1.32  & 87.8 & 13.98 & 5.88\\

    MediViSTA-SAM~\cite{kim2023medivista}  &   \ding{52} & \textbf{96.0}  &  \textbf{4.25} & \textbf{0.74} & \textbf{91.0} & \textbf{11.03} & \textbf{3.26} \\

    \hline
    \hline
    \multirow{2}{*}{Method} & \multirow{2}{*}{SAM-Based}  & \multicolumn{3}{c}{EndoNeRF} & \multicolumn{3}{c}{SuPer Data} \\
    
      & & PSNR  & SSIM  & LPIPS  & V1 & T1 & T2 \\
      
    \hline
    EndoNeRF~\cite{wang2022neural}  &  \ding{55}  & 21.4 & 0.72 &  0.29  & - & - & -\\

    SAMSNeRF~\cite{lou2023samsnerf}  &   \ding{52} & \textbf{34.5} & \textbf{0.92} & \textbf{0.10} & - & - & -\\ 
    
    DefSLAM~\cite{lamarca2020defslam}  &  \ding{55}  & - &- &  -  & 17.1 & 8.1 & 28.0 \\

    %SuPerPM-P~\cite{lin2023superpm}  &   \ding{52} & - &- & - & 11.1 & 7.2 & 43.4\\ 
    
    SuPerPM-F~\cite{lin2023superpm}  &   \ding{52} & - &- & - & \textbf{7.9} & \textbf{6.2} & \textbf{34.5} \\    
  
    \Xhline{1.2pt}  % 表格横线加粗
    \rowcolor[gray]{0.9} \multicolumn{8}{c}{\textbf{Optical Flow Estimation}} \\ 
    \hline 
    \multirow{2}{*}{Method} & \multirow{2}{*}{SAM-Based}  & \multicolumn{2}{c}{Sintel} & \multicolumn{2}{c}{Sintel Occ.}  & KITTI-15\\
    
      & & \multicolumn{1}{c}{clean} & \multicolumn{1}{c}{final} & \multicolumn{1}{c}{clean} & \multicolumn{1}{c}{final} & F1 \\
      
    \hline
    FlowFormer++~\cite{shi2023flowformer++}  &  \ding{55}  &1.07 & \textbf{1.94} &  6.64 & 10.63 & 4.52\\

    SAMFlow~\cite{zhou2023samflow}  &   \ding{52} &  \textbf{1.00} & 2.08 &  \textbf{5.97} & \textbf{10.60} & \textbf{4.49} \\

    \Xhline{1.2pt}  % 表格横线加粗
    \rowcolor[gray]{0.9} \multicolumn{8}{c}{\textbf{3D Point Cloud Segmentation}} \\ 
    \hline 
        \multirow{2}{*}{Method} & \multirow{2}{*}{SAM-Based}  & \multicolumn{5}{c}{ScanNet-2} \\
    
      & & wall & ﬂoor & cab & bed & Average\\
      
    \hline
    SparseConvNet~\cite{graham20183d}  &  \ding{55}  & \textbf{83.2} & \textbf{94.8} & \textbf{61.9} & \textbf{76.9} & \textbf{68.2}\\

    CSF~\cite{dong2023leveraging}  &   \ding{52} & 79.9 &87.8 &56.8& 65.2 & 65.1\\
    \Xhline{1.2pt}  % 表格横线加粗

  \end{tabular}
  }
  \label{tab:othervideounderstanding}

\end{table}

\subsection{Evaluation of Video Object Tracking Approaches}
\myPara{Datasets.} VOTS2023~\cite{kristan2023first} is the first dataset to merge short-term and long-term, as well as single-target and multi-target tracking, with dense mask annotations. It contains 144 videos and 341 targets in total. TREK-150~\cite{dunnhofer2023visual} is a dataset used to evaluate visual object tracking in the first person vision. It includes 150 videos with 97K densely annotated bounding boxes. NUT-L~\cite{yao2023sam} is a long-term nighttime UAV tracking benchmark consisting of 43 videos and 95,274 frames. YouTube-VOS 2018~\cite{xu2018youtube} is a large VOS dataset composed of 3,252 YouTube video clips and 133,886 object annotaions. It consists of 2,796 videos for the training set, 134 videos for the validation set, and 322 videos for the test set. Recently, VOS datasets such as YouTube-VOS 2018~\cite{xu2018youtube}, DAVIS 2016~\cite{perazzi2016benchmark}, and DAVIS 2017~\cite{pont20172017} are also adopted to evaluate VOT algorithms.

\myPara{Evaluation Metrics.} For the common VOT datasets (VOTS2023 and NUT-L), five popular evaluation metrics (\ie, success rate ($AUC$), accuracy ($A$), robustness ($R$), precision ($Pre$), and normalized precision ($nPre$))~\cite{kristan2023first,yao2023sam} are used. The one-pass evaluation ($OPE$), multi-start evaluation ($MSE$), and  human-object interaction evaluation ($HOI$) are adopted on the TREK-150~\cite{dunnhofer2023visual} dataset. For the YouTube-VOS 2018~\cite{xu2018youtube} dataset, the typical evaluation metrics include $\mathcal{J}$, $\mathcal{F}$, and $\mathcal{J\&F}$~\cite{perazzi2016benchmark,pont20172017}.

\myPara{Results Comparison.} Tab.~\ref{tab:videoobjecttracking} showcases the performance of eleven representative SAM-based methods and six current SOTA trackers on VOTS2023, TREK-150, NUT-L, DAVIS 2017 test, DAVIS 2016 val, and YouTube-VOS 2018 val.

We make the following observations: \textbf{1)} DMAOT~\cite{kristan2023first} achieves top performance with 63.6\% $AUC$ on the VOTS2023. This remarkable success can be attributed to the use of object-wise long term memory, which stores all masks of the tracked object in memory, and utilizes this memory to predict the current object mask, achieving more accurate results. \textbf{2)} Two SAM-based methods MSDeAOT~\cite{xu2023zju} and SAM-DA~\cite{yao2023sam} obtain the best results on the first person perspective tracking dataset TREK-150 and the long-term nighttime UAV tracking dataset NUT-L, respectively.  The former deploys transformers at multiple feature scales and converts bounding boxes to refined masks with the help of SAM, while the latter utilizes a SAM-powered target domain training sample swelling strategy to dispose domain adaptation. From the results of MSDeAOT and SAM-DA, we can observe that the powerful zero-shot generalization of SAM to generate precise masks and high-quality target domain training samples that are of great significance to specific video tasks. \textbf{3)} Using only click as initialization, the SAM-based method SAM-Track~\cite{cheng2023segment} scores 92.0\% and 79.2\% in terms of $\mathcal{J\&F}$ on DAVIS 2016 val and DAVIS 2017 test, which are comparable to the current SOTA tracker SwinB-DeAOT-L~\cite{yang2022decoupling}. \textbf{4)} SAM-based methods~\cite{chu2023zero,zhang2023personalize,rajivc2023segment,ke2023segment} lag far behind the current SOTA tracker (STCN~\cite{cheng2021rethinking}) on the YouTube-VOS 2018 dataset. This is because many SAM-based methods lack the space time correspondences module. By modeling spatiotemporal correspondences in the context of video frames, STCN reduces memory usage and more effectively utilizes information in memory and achieves top $\mathcal{J\&F}$ with 83.0\% on the YouTube-VOS 2018 val, which is significantly superior to its memory-based counterparts and SAM-based methods.

\subsection{Evaluation of other Video Understanding Approaches}

\myPara{Datasets.} FaceForensics++~\cite{rossler2018faceforensics} contains  1,004 videos, in which the forgery faces are generated by four deepfake algorithms (Deepfakes (DF), Face2Face (F2F), FaceSwap (FS), and NeuralTextures
(NT)). ViSha~\cite{chen2021triple} is a video shadow detection dataset, which comprises 120 videos, covering 7 shadow categories and 60 target categories with various lengths and different motion/lighting conditions. AVSBench~\cite{zhou2022audio} includes 4,932 videos, with two subsets: single source segmentation (V1S) and multiple sound source segmentation (V1M). Ref-DAVIS17~\cite{khoreva2019video} and Ref-Youtube-VOS~\cite{seo2020urvos} are used to evaluate RVOS. Medical video datasets include EndoVis2017~\cite{allan20192017}, EndoVis2018~\cite{allan20202018}, CAMUS~\cite{leclerc2019deep}, In-house Data~\cite{kim2023medivista}, EndoNeRF~\cite{wang2022neural}, and SuPer Data~\cite{lin2023superpm}. Sintel~\cite{butler2012naturalistic} and KITTI-15~\cite{geiger2013vision} are common datasets for optical flow estimation; ScanNet-2~\cite{dai2017scannet} is adopted in 3D point cloud segmentation.

\myPara{Evaluation Metrics.} In face forgery detection, accuracy is adopted for FaceForensics++ dataset~\cite{rossler2018faceforensics}. For video shadow detection, the common metrics include mean absolute error (MAE), F-measure (F$_{\beta}$), intersection over union
(IoU), and shadow/non-shadow region Balaence error rate (SBER/NBER)~\cite{chen2021triple}. Mean IoU (mIoU) and F-score are used on AVSBench~\cite{zhou2022audio}. Challenge IoU (cIoU), IoU, and mean class IoU (mcIoU)~\cite{yue2023surgicalsam} are adopted on EndoVis2017 and EndoVis2018. For the CAMUS dataset, common evaluation metrics include Dice, dH, and dA~\cite{kim2023medivista}. For the EndoNeRF dataset, PSNR, SSIM, LPIPS~\cite{lou2023samsnerf} are used. The reprojection errors~\cite{lin2023superpm} is adopted to evaluate algorithms on the SuPer Data-V1, T1, and T2. We report the clean/final EPE~\cite{zhou2023samflow} on Sintel test and Sintel test with occluded area, and F1 score on KITTI-15~\cite{geiger2013vision}. The mIoU metric is used for the ScanNet-2~\cite{dai2017scannet} dataset.

\myPara{Results Comparison.} Tab.~\ref{tab:othervideounderstanding} summarizes other video understanding results achieved by current SOTA and SAM-based methods. The main observations are presented as follows: \textbf{1)} Compared with VOS and VOT, SAM-based methods achieve more significant success in other video understanding tasks, \eg, deepfake detection, video shadow detection, RVOS, and various medical videos. One common feature of these tasks is that they often have small-scale datasets. These results confirm that developing a foundation model typically pre-trained on a broad dataset is a promissing direction to solve a range of downstream problems with limited data even distinct data distributions. \textbf{2)} For audio-visual segmentation, the carefully designed audio unmixing and semantic segmentation network (AUSS)~\cite{ling2023hear} achieves top results on the single-source subset, but the SAM-based method GAVS~\cite{wang2023prompting} is more competitive on the complex multi-source subset. GAVS is a simple encoder-prompt-decoder paradigm leveraging the prior knowledge of the visual foundation model SAM. This highlights the importance of utilizing the generalization ability of foundation models to solve specific tasks. \textbf{3)} Compared with fully supervised SparseConvNet~\cite{graham20183d}, the SAM-based method CSF~\cite{dong2023leveraging} achieves comparable results indicating that the effectiveness of adopting 2D foundation model to dispose 3D point cloud segmentation tasks. Inspired by the success of finetuning methods such as SuPerPM-F~\cite{lin2023superpm}, we believe that finetuning foundation models on downstream tasks is a promising direction for addressing huge domain gaps.

\begin{table*}[t]
  \centering

  %\captionsetup{font={small}}
  \caption{Performance evaluation of different video generation and video editing tasks. The best results are marked in \textbf{bold}.}

  %\footnotesize
  \setlength{\tabcolsep}{0.7mm}{
  \begin{tabular}{cccccccc} 

    \Xhline{1.2pt}  % 表格横线加粗
    \rowcolor[gray]{0.9} \multicolumn{8}{c}{\textbf{Video Synthesis}} \\ 
    \hline 
    
    \multirow{2}{*}{Method} & \multirow{2}{*}{SAM-Based}  & \multicolumn{6}{c}{Sequence of Pose Data~\cite{qin2023dancing}}\\
    
    & & Frame NIQE & Body NIQE & Background NIQE & Frame BRISQUE & Body BRISQUE &  Background BRISQUE \\
        
    \hline
    ControlVideo~\cite{zhang2023controlvideo}  &  \ding{55}  & 3.32 & 9.01 & 3.21 & 26.21 & 48.11 & 48.11\\

    Dancing Avatar~\cite{qin2023dancing}  &   \ding{52} & \textbf{2.99} & \textbf{5.03} & \textbf{2.44} &  \textbf{19.56} & \textbf{45.19} &  \textbf{43.75} \\

    \hline
    \hline 
    
    \multirow{2}{*}{Method} & \multirow{2}{*}{SAM-Based}  & \multicolumn{6}{c}{TikTok Dance Dataset~\cite{wang2023disco}}\\
    
    & & FID & SSIM  & PSNR & LISPIS  & FID-VID &  FVD \\
        
    \hline
    DreamPose~\cite{karras2023dreampose}  &  \ding{55}  & 72.62 & 0.511 & 28.11 & 0.442 & 78.77 & 551.02 \\

    DISCO~\cite{wang2023disco}  &   \ding{52} & \textbf{28.31} & \textbf{0.674} & \textbf{29.15} & \textbf{0.285} & \textbf{55.17} & \textbf{267.75} \\

    \Xhline{1.2pt}  % 表格横线加粗
    \rowcolor[gray]{0.9} \multicolumn{8}{c}{\textbf{Video Super-Resolution}} \\ 
    \hline 

    \multirow{2}{*}{Method} & \multirow{2}{*}{SAM-Based}  & \multicolumn{2}{c}{Vimeo-90K Fast~\cite{xue2019video}} & \multicolumn{2}{c}{Vimeo-90K Slow~\cite{xue2019video}} & \multicolumn{2}{c}{Vimeo-90K Average~\cite{xue2019video}}\\
    
    & & PSNR & SSIM  & PSNR & SSIM  & PSNR & SSIM \\
        
    \hline
    BasicVSR~\cite{chan2021basicvsr}  &  \ding{55}  & 38.2954 & 0.95152 & 32.5003 & 0.90629 & 35.3601 & 0.93287\\

    SEEM~\cite{Lu2023CanSB}  &   \ding{52} & \textbf{38.4074} & \textbf{0.95222}  & \textbf{32.6234} & \textbf{0.90775} & \textbf{35.4786} & \textbf{0.93390} \\

    \Xhline{1.2pt}  % 表格横线加粗
    \rowcolor[gray]{0.9} \multicolumn{8}{c}{\textbf{Video Editing}} \\ 
    \hline 

    \multirow{2}{*}{Method} & \multirow{2}{*}{SAM-Based}  & \multicolumn{2}{c}{Model Evaluation} & \multicolumn{3}{c}{User Study} \\
    
     &  & CLIP & DINO   & Quality  & Subject & Prompt  \\
        
    \hline
    DreamBooth-V~\cite{wu2023tune}  &  \ding{55}  & 0.301 &\textbf{0.509} &22.81 &24.69 & 15.00 \\

    Make-A-Protagonist~\cite{zhao2023make}  &   \ding{52} & \textbf{0.329} &0.457 & \textbf{67.50} & \textbf{63.44} & \textbf{66.25} \\

    \hline 
    \hline 

    \multirow{2}{*}{Method} & \multirow{2}{*}{SAM-Based}  & \multicolumn{6}{c}{TGVE~\cite{wu2023cvpr}} \\
    
    & & Text CLIPScore & Frame CLIPScore  & PickScore & Text Alignment & Structure & Quality \\
        
    \hline
    Text2Video-Zero~\cite{khachatryan2023text2video}  &  \ding{55}  & 25.88 & \textbf{92.07} & 19.82 & 0.448 & \textbf{0.493} & 0.516 \\

    2SVE~\cite{wu2023cvpr}  &   \ding{52} & \textbf{26.89} & 89.90 & \textbf{20.71} & \textbf{0.689} & 0.486 & \textbf{0.599} \\

    \Xhline{1.2pt}  % 表格横线加粗

  \end{tabular}
  }
  \label{tab:videogenerationandediting}

\end{table*}

\subsection{Evaluation of Video Generation and Editing Approaches} 
\myPara{Video Generation.} We first summarize video synthesis results on Sequence of Pose Data~\cite{qin2023dancing} and TikTok Dance~\cite{wang2023disco} datasets. The former contains 10 high-resolution human dance videos, while the later includes around 350 videos with video length of 10 to 15 seconds. For the Sequence of Pose Data, BRISQUE and NIQE~\cite{qin2023dancing} are introduced to assess the frame quality. We report frame-wise metrics (FID, SSIM, LISPIS, and PSNR), and video-wise metrics (FID-VID and FVD)~\cite{wang2023disco} on the TikTok Dance dataset. Tab.~\ref{tab:videogenerationandediting} presents the video generation quality evaluation of SAM-based methods (Dancing Avatar~\cite{qin2023dancing}, DISCO~\cite{wang2023disco}) and current SOTA methods (ControlVideo~\cite{zhang2023controlvideo} and DreamPose~\cite{karras2023dreampose}). It is evident that SAM-based methods exhibit a significant superiority compared to current SOTA methods in video generation tasks. Similar results can be observed from the VSR dataset Vimeo90K~\cite{xue2019video}, and the SAM-based method (SEEM~\cite{Lu2023CanSB}) significantly surpasses the current SOTA method (BasicVSR~\cite{chan2021basicvsr}).

\myPara{Video Editing.} Tab.~\ref{tab:videogenerationandediting} compares the results of generic video editing of the SAM-based method (Make-A-Protagonist~\cite{zhao2023make}) and the current SOTA method (DreamBooth-V~\cite{wu2023tune}), using model evaluation (CLIP~\cite{radford2021learning} and DINO~\cite{caron2021emerging}) and user study (quality, subject, and prompt). Make-A-Protagonist achieves better model evaluation scores and an overwhelming preference in terms of user study. Furthermore, the SAM-based method (2SVE~\cite{wu2023cvpr}) is significantly better than its counterpart (Text2Video-Zero~\cite{khachatryan2023text2video}) on the challenging text guided video editing dataset TGVE~\cite{wu2023cvpr}. 

Overall, the above results demonstrate that the visual foundation model SAM exhibits exceptional performance in video generation and editing tasks, although it is essentially for image segmentation. This recent trend of \emph{SAM for anything}, has resulted in excellent performance in many other domains such as non-Euclidean domain~\cite{Non-euclidean_segment} and adversarial attacks~\cite{zhang2023attack}, paving the way to explore task-agnostic foundation models for vision and beyond~\cite{chunhui2023samsurvey}.

%------------------------------------------------------------------------------
\section{Conclusion and Future Directions}
\label{sec:Outlooks}

\subsection{Conclusion} This survey offered an in-depth look at the latest developments in the era of foundation models with a focus on SAM for videos. To the best of our knowledge, this is the first systematic and comprehensive survey that concentrate on this specific and promising research field. We commenced by summarizing the  unique challenges in the video domain, highlighting the extreme complexity of video tasks and the urgent need for a systematic review of SAM models for videos. This was followed by an overview of SAM and SAM 2, different research routes, and video-related research domains. Building upon above foundation, we exhaustively reviewed existing works and divided them into three key areas: video understanding, video generation, and video editing, considering their distinct technical perspectives and research objectives. In addition, we provided comparative results of SAM-based methods and current SOTA methods on various video tasks, together with numerous insightful observations. 

\subsection{Future Directions} Through our investigation and in-depth evaluation, we have found that although the SAM models (including SAM 2) has made or is making significant breakthroughs in various image and video tasks, there still exist numerous opportunities and challenges. We provide several future research directions in the area of SAM for videos and beyond in the following.

\myPara{$\bullet$ Constructing Large-Scale Video Datasets.} The substantial achievements of visual foundation models are mainly attributed to the availability of billions of high-quality image data. Nevertheless, considering the huge cost of data collection and annotation, current video tasks are usually limited to relatively small-scale datasets. For instance, the VOT dataset TrackingNet~\cite{muller2018trackingnet} contains 30,643 videos and 14.43 million frames, but its significant drawback is sparse annotation. Leveraging SAM to automatically generate dense mask annotations from videos is a potential solution to achieve data  scalability~\cite{videoTextSpottingSA}.

\myPara{$\bullet$ Building Large-Scale Video Foundation Models.} Most current visual foundation models primarily concentrate on pretraining and adaptation at the image level, which are evidently constrained in complex and dynamic video-level understanding tasks. Due to the increasingly convenient collection and storage, videos are emerging as a domain force on the edge devices and Internet~\cite{xing2023videodiffusion}. Therefore, the development of video foundation models, \eg, medical video foundation models, for broad video applications becomes an urgent requirement.

\myPara{$\bullet$ Parameter-Efficient Training and Fast Inference.} Training video foundation models with billions of parameters from scratch inevitably faces significant challenges due to high data dimension and the high computational overhead. While some efforts to explore new technologies, \eg, adapter~\cite{chen2023sam} and prompt learning~\cite{li2023auto}, by utilizing pre-trained models to promote efficient transfer learning, there remains a pressing need to mitigate training and inference expenses. More efficient training  strategies and model compression methods may unlock more power in video foundation models on edge devices, \eg, automobile and surgical robots, with limited computational resources.

\myPara{$\bullet$ Incorporating More Modalities.} Although current foundation models have achieved significant advances in single modality and two modalities (\eg, vision and text, vision and audio), the integration of more modalities is far from being explored. A core reason is the lack of extensive aligned multi-modal data~\cite{wang2023large}. On one hand, collecting multi-modal data, \eg, visual images, text, audio, point cloud, infrared images, depth images, and event streams, is crucial for researching multi-modal foundation models. On the other hand, developing a unified model~\cite{zhang2023meta} for multi-modal perception without requiring paired multi-modal data is a promising direction.

\myPara{$\bullet$ Credible and Interpretable Video Foundation Models.} The security of artificial intelligence has attracted significant concerns as it may lead to privacy breaches and security risks in practical applications such as face recognition and autonomous driving. However, the capability of video foundation models to resist various attacks~\cite{zhang2023attack} is still far from being explored. In addition, due to the high complexity and rapidly increasing deployment of video foundation models~\cite{chunhui2023samsurvey}, improving their interpretability and enhancing people's trust in decision-making is a valuable avenue for future research.

\myPara{$\bullet$ More Innovative Opportunities in SAM for Videos.} 
As SAM for videos is a rapid-evolving research field, we might not cover all the latest advancements in this review. Actually, there are masses of video tasks that are not covered by SAM or have not been fully studied, such as video captioning, video-based event detection, video-based behavior/action recognition, video summarization, and video frame interpolation. Last but not least, combining SAM with a large number of traditional technologies/methods (\eg, knowledge distillation, and graph learning) and cutting-edge technologies/methods (\eg, video diffusion model, explainable AI (XAI), and embodied AI) can stimulate more opportunities in the era of foundation models due to SAM's versatility and plug-and-play nature.

%------------------------------------------------------------------------------
\myPara{Acknowledgement.} This work was supported by the National Natural Science Foundation of China (No. 62101351), and the Key Research and Development Program of Chongqing (cstc2021jscx-gksbX0032).

%%%%%%%%% REFERENCES
\bibliographystyle{IEEEtran}
\bibliography{ref}
%\bibliography{ref_short} % for submitted

\end{document}